\documentclass[10pt]{article} 

\usepackage[accepted]{rlj} 

\usepackage{amssymb}            
\usepackage{mathtools}          
\usepackage{mathrsfs}           
\usepackage{graphicx}           
\usepackage{subcaption}         
\usepackage[space]{grffile}     
\usepackage{url}                

\usepackage{amsfonts}           
\usepackage{nicefrac}           
\usepackage{microtype}          
\usepackage{booktabs}           
\usepackage{makecell}           
\usepackage{pifont}             
\usepackage{placeins}           
\usepackage{wrapfig}            
\usepackage{float}              
\tcbuselibrary{most}            
\usepackage{xcolor}

\newcommand{\cmark}{\ding{51}}
\newcommand{\xmark}{\ding{55}}

\graphicspath{ {./images/} }

\title{When are LLMs Sufficient Policy Optimizers for Sequential RL Tasks?}

\setrunningtitle{When are LLMs Sufficient Policy Optimizers?}

\author{Stephane Hatgis-Kessell, Emma Brunskill}

\emails{stephhk@stanford.edu}

\affiliations{
\textbf{Department of Computer Science, Stanford University}
}

\begin{document}

\maketitle

\begin{abstract}
We study when large language models (LLMs) can serve as effective black-box policy optimizers for reinforcement learning (RL) tasks, i.e., when can we replace classical RL algorithms with an LLM? We explore this question by introducing Prompted Policy Optimization (PromptPO), an iterative method that prompts an LLM with Python descriptions of the state space, action space, and reward function, then has it generate and refine executable policies based on rollout feedback. Across hard exploration environments, Meta-World robotics tasks, and several real-world control problems, PromptPO often matches or exceeds the performance of standard RL baselines while using substantially fewer environment interactions. To maximize expected return, and without further explicit prompting, the policies PromptPO outputs range from tuned proportional controllers or rule-based plans to policies that run planning algorithms like value iteration. Our results demonstrate that LLM-based policy optimization is sufficient when the LLM can leverage prior knowledge about the environment or optimization strategy. PromptPO underperforms standard RL baselines in MuJoCo domains. This demonstrates possible limitations of LLM-based policy optimization to settings that requiring fine-grained continuous control. 
\end{abstract}

\section{Introduction}
\begin{wrapfigure}{r}{0.5\textwidth}
    \vspace{-20mm}
    \centering
    \includegraphics[width=0.48\textwidth]{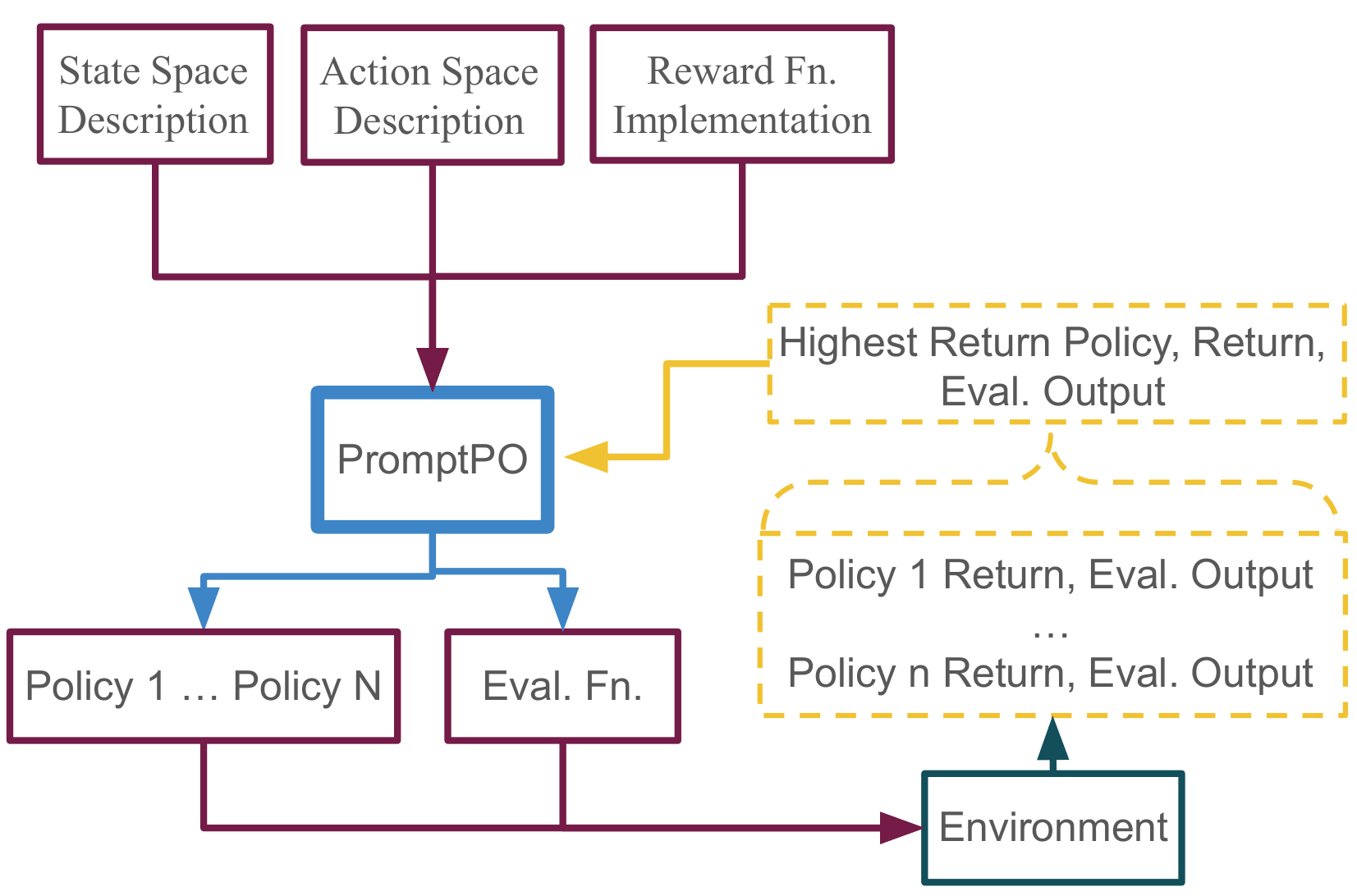}
    \caption{PromptPO input: a description of the state space, action space, and reward function in Python code. We avoid inputting context about the environment's transition dynamics to evaluate PromptPO in model free settings. PromptPO generates a set of policies and an evaluation function, both implemented in Python code. The policies are rolled out in the environment, evaluated with respect to the evaluation function and reward function, and then the best performing policy is fed back to PromptPO. PromptPO proposes a new policy at the next round to maximize expected return.}
    \label{fig:main_fig}
\end{wrapfigure}
A central goal of reinforcement learning (RL) is to produce a policy that maximizes expected return with respect to a reward function \citep{sutton2018reinforcement}. To this end, an influential line of RL research has focused on developing algorithms to train a policy that maximized expected return through continued interaction with an environment \citep{watkins1992q,sutton2000policy,mnih2015human,schulman2017proximal,haarnoja2018soft,schulman2015trust,fujimoto2018addressing}. These methods are often computationally intensive---requiring substantial interaction with the environment---and brittle, as they depend on careful hyperparameter tuning and the selection of an appropriate RL algorithm. The choice of hyperparameters, model architecture, RL optimization algorithm, and other implementation or methodological details can significantly influence the performance of the learned policy \citep{henderson2018deep,andrychowicz2021what,eimer2023hyperparameters}. Even though RL has become widely used, it continues to be quite challenging in practice to get right in many settings, especially in sequential decision-making settings. (See \citet{li2022reinforcement,dulacarnold2019challenges} for a discussion attempting to answer: ``Why has RL not been widely adopted in practice yet?'')

In this paper we investigate when Large Language Models (LLMs) can produce a policy that maximizes expected return as a drop in replacement for standard RL algorithms that doesn't require a long list of design decisions to get right (e.g., choice of model architecture, RL algorithm, hyperparameters) and that is potentially more sample efficient. More broadly, we argue that LLM-based optimization can serve as a competitive alternative to standard RL algorithms for many canonical sequential decision making RL tasks.

Our claim is based on our empirical evidence of  prompting an LLM\footnote{For this work, we use Gemini 3 Pro.} to return a policy that maximizes expected return, given a Python-formatted description of the state space, action space, and reward function. To this end we introduce Prompted Policy Optimization (PromptPO), outlined in Figure \ref{fig:main_fig}. PromptPO, perhaps surprisingly, matches or exceeds the performance of the standard RL algorithms like SAC \citep{haarnoja2018soft}, PPO \citep{schulman2017proximal}, and DQN \citep{mnih2015human}, on several delayed-reward environments that require substantial exploration, including gridworld MDPs with randomly generated transition dynamics and Point Maze, a challenging exploration environment. It is also significantly more sample efficient on multiple robotic manipulation tasks from Metaworld, and in three real-world control tasks.   PromptPO underperforms standard RL algorithms on a suite of MuJoCo continuous control tasks, indicating that current LLMs may have difficulty designing policies for environments that require fine-grained control.

Interestingly, to maximize expected return and without further explicit prompting, the policies PromptPO outputs range from tuned proportional controllers or rule-based plans to policies that run planning algorithms like value iteration. Broadly, we show that LLMs are sufficient black box policy optimizers for many environments. In the rest of the paper we describe our approach, then experiments, before concluding with implications and next steps.

We release our code base \href{https://github.com/Stephanehk/PromptPO}{here}, including the minimal implementation of PromptPO used for this work, \href{https://github.com/Stephanehk/PromptPO/tree/main/env_contexts}{the context we provided to PromptPO for each environment}, and \href{https://github.com/Stephanehk/PromptPO/tree/main/representitive_policies}{examples of the policies PromptPO generates} for each environment.

\section{Related Works}

We follow the same procedure as a line of prior work that prompts an LLM to iteratively generate a code artifact \citep{karpathy2026autoresearch, hennes2026code, novikov2025alphaevolve}  or solution \citep{yuksekgonul2024textgrad, lee2025feedback,agrawal2025gepa} to optimize for a specific objective. We investigate when this framework is sufficient to solve classical, single agent sequential decision making tasks as a substitute for standard RL algorithms, and in particular how this framework compares to standard RL algorithms with respect to environment sample efficiency. Following the insights from \citet{liang2023code} that policies can be represented as executable programs, we represent generated policies as Python code. Our work differs from \citet{liang2023code} and a line of others \citep{wang2023voyager, ahn2022can, huang2022inner, shinn2023reflexion, huang2022language} in that they use LLMs to generate policies to satisfy a natural language description of a task or reward function. Rather, we study the setting where an LLM must produce a policy to optimize expected return with respect to a specified reward function, i.e., the standard RL setting.  We summarize the differences between our work and these in Table \ref{tab:llm_methods_comparison}.

Other work focuses more directly on using LLMs for policy optimization, such as prompting an LLM to output an action directly at every decision making step \citep{brooks2023large, szot2023large}. This approach doesn't naturally scale to the long-horizon decision making settings we consider. \citet{zhou2025prompted} uses an LLM to generate policy parameter vectors, but they assume the policy can be expressed with a small, human-specified set of parameters that the LLM can reliably output, limiting expressivity and requiring manual design of the policy class. We investigate the more general setting where an LLM must output the policy itself, not parameters for a prespecified policy class.

\begin{table}[t]
\centering
\scriptsize
\setlength{\tabcolsep}{3pt}
\renewcommand{\arraystretch}{0.95}
\caption{Comparison of LLM-based optimization methods across key properties. To evaluate LLM-based optimization as a drop-in-replacement for RL algorithms, PromptPO focus on all properties.}
\label{tab:llm_methods_comparison}
\begin{tabular}{lccc}
\toprule
\textbf{Method} &
\makecell{\textbf{Evaluates in sequential}\\\textbf{decision making}} &
\makecell{\textbf{Uses specific}\\\textbf{optimization target}} &
\makecell{\textbf{Evaluates for sample}\\\textbf{efficiency improvements}} \\
\midrule

AutoResearch~\citep{karpathy2026autoresearch}
& \xmark & \cmark & \xmark \\

GEPA~\citep{agrawal2025gepa}
& \xmark & \cmark & \cmark \\

AlphaEvolve~\citep{novikov2025alphaevolve}
& \xmark & \cmark & \xmark \\

TextGrad~\citep{yuksekgonul2024textgrad}
& \xmark & \cmark & \xmark \\

Feedback Descent~\citep{lee2025feedback}
& \xmark & \cmark & \cmark \\

Voyager~\citep{wang2023voyager}
& \cmark & \xmark & \cmark \\

Can~\citep{ahn2022can}
& \cmark & \xmark & \xmark \\

Inner Monologue~\citep{huang2022inner}
& \cmark & \xmark & \xmark \\

Reflexion~\citep{shinn2023reflexion}
& \cmark & \xmark & \xmark \\

Language Models as Agents~\citep{huang2022language}
& \cmark & \xmark & \xmark \\

CSRO~\citep{hennes2026code}
& \cmark & \cmark & \xmark \\

\textbf{PromptPO (ours)}
& \textbf{\cmark} & \textbf{\cmark} & \textbf{\cmark} \\

\bottomrule
\end{tabular}
\end{table}

\section{Prompted Policy Optimization (PromptPO)}


PromptPO inputs a description of the state space, action space, and the reward function implementation. These inputs are expressed in Python code, and naturally accompany environments designed for RL such as via Gymnasium \citep{towers2023gymnasium} interfaces. Importantly, we avoid providing context about the environments' transition dynamics to understand PromptPO's ability to maximize expected return without explicit model based knowledge. All context used by PromptPO is available in our Github repository.

We first ask PromptPO to implement an evaluation class called Feedback via the prompt summarized in Figure \ref{fig:feedback_prompt}. The implemented evaluation provides additional supervision beyond the environment reward function. Common RL algorithms reason about prior interactions by way of Bellman backups or temporal difference errors propagated through the state-action space. In contrast, similar to REINFORCE, PromptPO updates policies using only trajectory-level feedback, making the design of evaluation functions beyond the reward signal a key component of the method. While future work could incorporate state--action level feedback into PromptPO, we hypothesize that current LLMs may struggle to reason over long-horizon sequences of individual transitions, particularly in environments with hundreds or thousands of steps. Upon generating an evaluation function, PromptPO is then prompted to generate a policy to maximize expected return using a prompt summarized in Figure \ref{fig:policy_generation_prompt}.

We sample $N$ candidate policies from PromptPO and roll out each policy in the environment. For each rollout, we compute the return under the provided reward function and evaluate performance using the generated evaluation function. The best-performing policy---measured by return---is then reported back to PromptPO, which produces a natural language reflection on its performance using the prompt in Figure~\ref{fig:policy_evaluation_prompt}. The history of the best policy from all previous rounds is maintained in context. Finally, this reflection, along with the history of prior best policies and their associated reflections, is used to generate a new policy via the prompt in Figure~\ref{fig:policy_generation_prompt}. At any point in the iterative process, if a Python execution error occurs due to incorrectly generated code (e.g., in the policy or evaluation), PromptPO is provided with the error and prompted to re-implement the relevant code artifact.

\section{Experiments}
\label{sec:exp_summary}
\begin{figure}[t]
    \centering
    \includegraphics[width=0.8\linewidth]{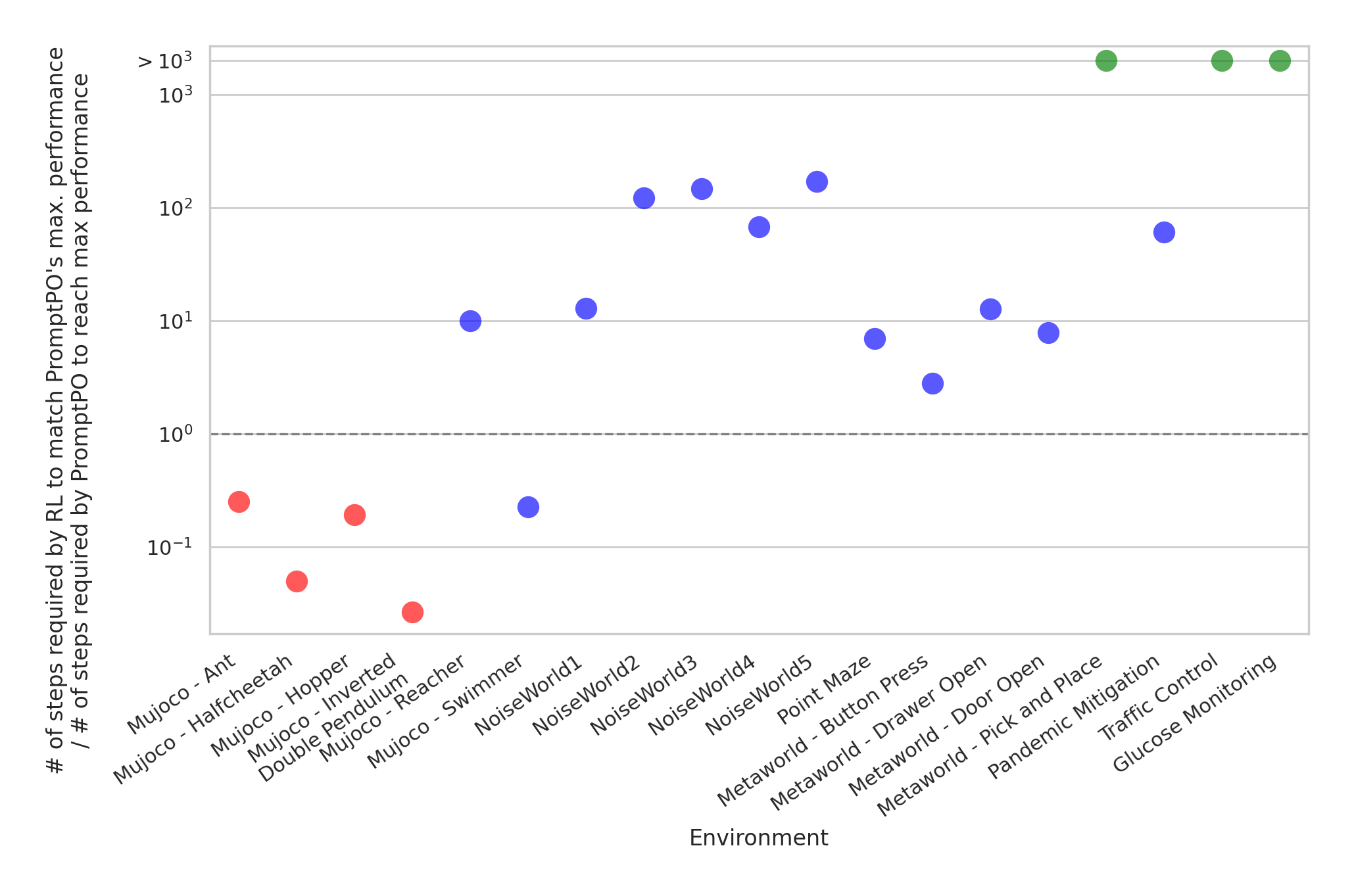}
    \caption{\textbf{Comparison of PromptPO to best performing RL algorithm in terms of final performance (color) and sample efficiency (y position).} Green points are environments where PromptPO attains a higher mean return than RL. Blue points are environments where PromptPO attains the same mean return as RL, and red points are those where it attains a lower mean return. All points above the gray dotted line are environments where PromptPO is more sample efficient, and the y-axis denotes by how much. Mean return is computed over 3 seeds. The plotted RL performance corresponds to the best-performing algorithm for each environment, selected from the suite of RL methods described in Section~\ref{sec:detailed_desc}.}
    \label{fig:sample_efficiency_summary}
    \vspace{-5mm}
\end{figure}

\begin{figure}[t]
    \centering
    \includegraphics[width=0.8\linewidth]{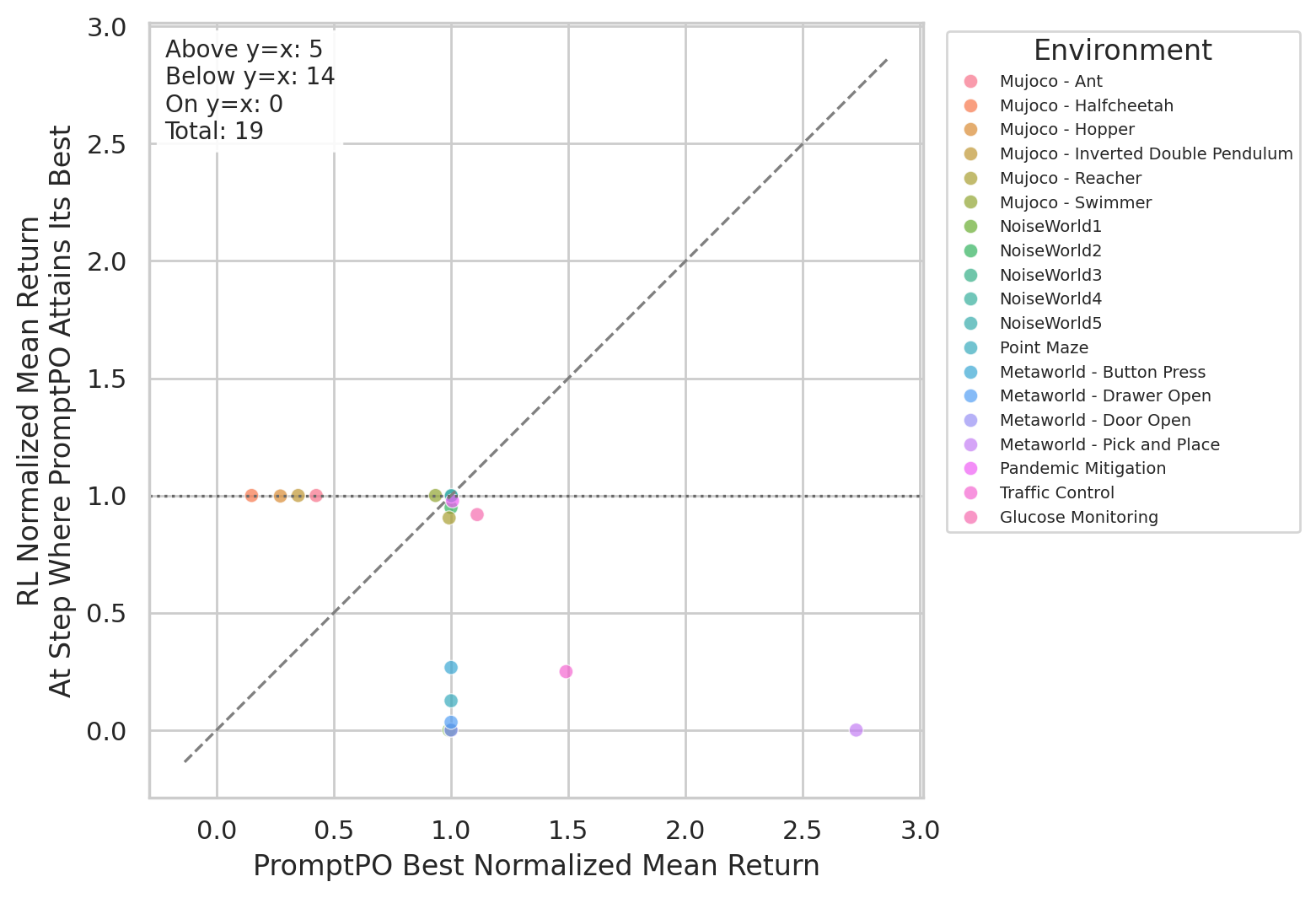}
    \caption{\textbf{Comparison of PromptPO to best performing RL at the step when PromptPO achieves its best performance}. Returns are normalized such that a uniformly random policy has value 0 and the best-performing RL policy has value 1; values greater than 1 indicate that PromptPO outperforms RL's best policy. Points below the line $y = x$ correspond to environments where PromptPO attains higher performance than RL at the time PromptPO reaches its best performance. All other details follow Figure \ref{fig:main_fig}.}
    \vspace{-5mm}
    \label{fig:norm_mean_ret_summary}
\end{figure}

We evaluate PromptPO against a suite of standard RL algorithms across four classes of domains: hard-exploration environments, robotic manipulation and continuous control, and real-world control tasks. Our RL baselines span the most widely used model-free algorithms: PPO \cite{schulman2017proximal}, SAC \cite{haarnoja2018soft}, DQN/QR-DQN \cite{mnih2015human}, and TD3 \cite{fujimoto2018addressing}, along with variants augmented with hindsight experience replay \citep{andrychowicz2017hindsight} where it aids exploration. For each environment, we report the mean return of the best-performing RL baseline. The domains we test on are intentionally diverse — ranging from randomly generated gridworlds and maze navigation, to MuJoCo and Meta-World robotics, to three real-world control problems — and are chosen to probe both where LLM-based policy optimization succeeds and where it falls short.

Figure \ref{fig:sample_efficiency_summary} and \ref{fig:norm_mean_ret_summary} illustrate a summary of our results when $N=10$ policy candidates are sampled from PromptPO at each round. PromptPO matches or outperforms the best performing RL algorithms in 15/19 environments, is substantially more sample-efficient in 14/19 environments, and is more than an order of magnitude more sample efficient in 11/19 environments. Section \ref{sec:detailed_desc} adds detail to the content below, including environment specific results. 

\textbf{Exploration Environments~~} We introduce NoiseWorld, a suite of randomly generated grid worlds designed to evaluate PromptPO's exploration capabilities while mitigating the LLM's prior knowledge of the environment. Prior work has taken a similar approach, generating random environments in other settings \citep{duan2016rl}. We additionally evaluate PromptPO on a physics-based maze navigation domain---the Point Maze Gymnasium environment \citep{gymnasium2023}---using a custom maze layout. To construct the RL baselines for NoiseWorld, we evaluate PPO, QR-DQN, and variants of both with hindsight experience replay \citep{andrychowicz2017hindsight}---which often aids exploration---and report the mean return of the best performing method for each board layout. For Point Maze, we consider SAC, PPO, and variants of both with hindsight experience replay, reporting the mean return of the best performing method. See Appendix \ref{app:noiseworld_hyperparams} for details on our choice of hyperparameters and implementation details for NoiseWorld. For Point Maze, we use the hyperparameters from \citet{liu2025courage}.

Across these environments, PromptPO matches the performance of the best-performing RL algorithms, demonstrating strong exploration abilities driven by sampling and evaluating many candidate policies at each update round. Notably, 2/5 of the NoiseWorld environments remain unsolved by both PromptPO and RL baselines. Further details are provided in Section~\ref{sec:exploration_env}. Interestingly, in the NoiseWorld environments, PromptPO's generated policies implement a planning algorithm, such as Dijkstra's algorithm or Value Iteration using the board layout, which is provided in the observation, together with an empirically estimated transition model derived from prior experience. The resulting planning algorithm is implemented directly in the Python policy class and is executed at every environment timestep. This behavior indicates the effectiveness of PromptPO's prior on what optimization procedure to employ. In Point Maze, PromptPO's implemented policy is a proportional controller.

\textbf{Robotics Environments~~} We evaluate PromptPO on six MuJoCo~\citep{todorov2012mujoco} and four Meta-World robotic manipulation environments~\citep{yu2020metaworld}. A key distinction is that MuJoCo tasks require actions in the form of joint torques, whereas Meta-World tasks operate over end-effector displacements. For each Mujoco environments, to construct our standard RL baseline, we train a policy with PPO, SAC, and TD3 and report the mean return of the best performing method per environment. For Metaworld, we train a policy with SAC which \cite{yu2020metaworld} demonstrates dominates alternative RL algorithms. We use the hyperparameters from \citet{raffin2021rlzoo} for the Mujoco environments, and from \citet{yu2020metaworld} for Metaworld.

PromptPO underperforms standard RL algorithms on the MuJoCo suite, but in Meta-World it achieves greater sample efficiency in all tasks and improves final performance on one task. These results highlight a limitation of PromptPO: it struggles with fine-grained continuous control (e.g., joint torques), but performs well when the action space is more amenable to natural language reasoning. See Section~\ref{sec:robotics_envs} for additional details. Across all robotics environments, PromptPO generates proportional controllers and tunes their parameters as it receives additional environment feedback.

\textbf{Real World Control Environments~~} We evaluate PromptPO on three real-world decision-making environments used by \citet{pan2022effects}: administering insulin to patients with Type II diabetes, determining lockdown regulations to manage a Covid-19 pandemic, and controlling the accelerations of a fleet of autonomous vehicles merging onto a highway. To construct the standard RL baseline, we train a policy with PPO in all environments, following the implementation and hyperparameters of \citet{laidlaw2403correlated} for these environments.

PromptPO is more than $60\times$ more sample efficient in the Pandemic Mitigation environment, and outperforms the final performance of PPO for the training budget we consider in the Glucose Monitoring and Traffic Control environments. These results suggest that PromptPO can be a strong policy optimizer in real-world settings, where the LLM may leverage prior knowledge about the environment from pretraining to generate effective policies. See Section~\ref{sec:real_world_envs} for additional details. In these environments, PromptPO implements rule-based heuristic policies that are iteratively refined as additional environment feedback is obtained.

\textbf{On Pretraining Biases and Privileged Information in PromptPO~~} Most of the environments we consider are publicly available and likely present in the LLM's pretraining data. As a result, PromptPO may benefit from strong inductive biases or access to information about environment dynamics or effective policies that would not typically be available to an RL agent. We take several steps to mitigate these concerns. First, we evaluate PromptPO on newly generated environments (e.g., NoiseWorld) and novel configurations of existing environments (e.g., Point Maze) that are unlikely to appear in pretraining data. We do not allow access to external tools such as internet search, and we do not provide transition dynamics in context. We additionally audit generated policies using LLM-guided search to check for reliance on publicly available solutions or privileged information. In one surfaced instance, for example, PromptPO reproduced a maze layout from publicly available documentation; we mitigate this by using a custom layout. While prior exposure to related environments likely provide useful inductive biases in ways that are hard to predict, PromptPO must still synthesize policies that maximize expected return in the specific instances we consider.

\textbf{On Comparing PromptPO to RL from Scratch~~}
We compare PromptPO to RL methods that optimize for expected return starting from a randomly initialized policy. This comparison is intentionally asymmetric: while RL methods learn solely from environment interaction, PromptPO leverages substantial prior knowledge from pretraining. Our goal is not to equalize these settings, but to understand the extent to which pretraining can substitute for environment interaction in canonical RL tasks. In particular, this comparison targets a central question: how much useful structure about policies and environments is already captured by large-scale pretraining, and how effectively can it be leveraged for decision making? From a practical perspective, we argue that such comparisons are appropriate, as practitioners can choose between methods that learn from scratch and, via LLM based optimization methods, those that use pretrained models. 

\section{On the Limitations of PromptPO}
PromptPO performs poorly on MuJoCo environments, which require fine-grained continuous control. This result is in line with prior work that similarly finds LLMs struggle to comprehend low-level robotic commands such as joint torques \citep{tang2023saytap}. We expect similar limitations in other environments with comparable action spaces. However, these environments can sometimes be reformulated to better suit PromptPO---for example, by redefining the action space. Indeed, PromptPO performs well in MetaWorld, where actions correspond to end-effector displacements rather than low-level motor torques. We also suspect that PromptPO will perform poorly in environments where it cannot reason over the state space, action space, or reward function in natural language. For example, if the reward function were instead parameterized as a neural network---perhaps learned via RLHF---then PromptPO may struggle to generate a policy that maximizes expected return.

Interestingly, PromptPO matches or exceeds RL performance even in environments where it likely lacks environment-specific priors---such as NoiseWorld---suggesting it may generalize well to other unfamiliar environments. This performance may stem not from prior knowledge of the environment itself, but from a prior over effective optimization strategies.

\section{Discussion and Conclusion}

Our results suggest that LLM-based optimization methods like PromptPO can offer a compelling alternative due to their strong priors and resulting sample efficiency. PromptPO is easy to use—it requires few algorithmic hyperparameters to tune and interfaces with natural-language information a practitioner can readily provide—and proved to be a strong policy optimizer across many of the environments we considered: random grid worlds, maze navigation, Meta-World robotics, pandemic mitigation, glucose monitoring, and autonomous driving. Practitioners without extensive RL expertise may find PromptPO easier to apply to sequential decision-making problems than standard RL methods. 

The sample efficiency improvements of PromptPO highlight the benefit of leveraging LLM methods that are built on large-scale pretraining. While much of RL research focuses on training policies from scratch, leveraging the rich priors and knowledge embedded in LLMs can serve as a significant advantage. There has been a long debate (see Richard Sutton's bitter lesson \cite{sutton2019bitter}) over the strengths and weaknesses of injecting human knowledge into AI and RL algorithms. Methods such as Thompson Sampling and many others can leverage prior structure or information to shape and speed learning. LLMs are built on far more expansive data and knowledge, and pure LLM optimization for policies may offer practical advantage beyond single-human-curated injected priors and intuition for RL algorithm selection. While the human inductive biases embedded in LLMs may limit PromptPO methods from significantly surpassing human capabilities, PromptPO could still be a useful approach for generating high-performing policies across many environments and may facilitate broader adoption of RL in real-world settings---particularly among practitioners without RL expertise.

PromptPO could also be considered an algorithm to evaluate within future RL research.  For many research questions and methods, particularly those used in conjunction with existing policy optimizers---such as to encourage exploration, abide by safety constraints, leverage world models for planning, or enable continual learning---researchers often show that their proposed method induce performance gains across different RL algorithms. Additionally showing performance gains with LLM-based policy optimization methods would be a practical contribution. One challenge for using PromptPO in research evaluations is that LLMs may have already been trained on many existing RL tasks and benchmarks, and so such models may have "peaked at the test set" in producing new policies. For this reason, environments with randomly generated transition dynamics could provide a useful testbed for LLM-based policy optimization methods.

Finally, we observe that in the NoiseWorld environments, PromptPO implements Value Iteration without being explicitly prompted to do so. In the continuous control environments, like in Mujoco or Metaworld environments, PromptPO implements and tunes a proportional controller. This provides evidence that PromptPO can select appropriate policy optimization procedures for a given setting, suggesting that future LLM-based optimization methods may similarly implement more sophisticated existing RL algorithms---or even develop novel ones that extend beyond current approaches.

\vspace{-4mm}


\bibliography{main}
\bibliographystyle{rlj}

\beginSupplementaryMaterials

\section{Detailed Environment Descriptions and Experiments}
\label{sec:detailed_desc}

We provide additional details about the set of environments we use for evaluation in Section \ref{sec:exp_summary}, and present results for each individual environment. In Appendix~\ref{app:varrying_n}, we present results for PromptPO trained with different values of $N$, the number of candidate policies generated per round.

\subsection{Exploration Environments}
\label{sec:exploration_env}
\textbf{NoiseWorld~~} We propose NoiseWorld to evaluate PromptPO in a setting where it doesn't reasonably have access to a strong prior on the environment from its training data. NosiseWorld is a grid-world environment where the agent starts in the top left corner and must maximize its expected return by navigating to a goal state in the bottom right corner. The action space consists of moving in the 4 cardinal directions. By default, each NoiseWorld board is $10\times10$ with a horizon of $100$. The agent accrues a $-1$ reward at each timestep, and $+1000$ reward for reaching the goal state. NosieWorld has additional 6 cell types:
\begin{itemize}
    \item Cell type 0 is a blank; transitions out of cell type 0 are deterministic
    \item Cell type 1 is a wall; no transitions into these cells are successful.
    \item Cell type 2 is such that any transitions out of this cell are successful with probability $p$, and otherwise result in the agent staying in the current cell with probability $(1-p)$.
    \item Cell type 3 is such that any transitions out of this cell are successful with probability $p$, and otherwise result in the agent transitioning back to the start state with probability $(1-p)$.
    \item Cell type 4 ends the episode with a penalty of $-1000$.
\end{itemize}

We instantiate NoiseWorld1, NoiseWorld2, and NoiseWorld3 where cell types are sampled from the types listed above, and where relevant, the transition success probabilities are also sampled for each cell. When providing the observation space to PromptPO, which consists of the agents current location and the board layout, we do not indicate which cells types correspond to what transition dynamics or behaviors; PromptPO must figure this out on its own.

We additionally evaluate PromptPO with NoiseWorld4 and NoiseWorld5, which in addition to the cell types above, also have cell types 5 and 6; there is exactly one of these cell types each per board, and to attain the reward of $+1000$ upon visiting the goal state, the agent must visit the cell of type 5 and then the cell of type 6 first. Otherwise, reaching the goal state ends the episode with no bonus.

\textbf{Point Maze~~} We evaluate PromptPO in the Point Maze environment \citep{farama_pointmaze}, where a 2-DoF ball that is force-actuated must navigate through a maze to reach a target goal. The action space is continuous and 2-dimensional, and the observation consists of the balls position and velocity, the goal's position, and whether the goal has been reached. Each episode has a horizon of 800, and a delayed reward of +1 for reaching the goal and 0 otherwise. We generate a new $10 \times 9$ Point Maze map that differs from the default map provided by \citet{farama_pointmaze} to force PromptPO to explore the environment rather than rely on the LLM's training data. We found that when we did not do this, PromptPO---without access to the internet---recalled the default map used by Point Maze Large from their documentation.

\begin{figure}[t]
    \centering
    \begin{subfigure}{0.19\textwidth}
        \centering
        \includegraphics[width=\linewidth]{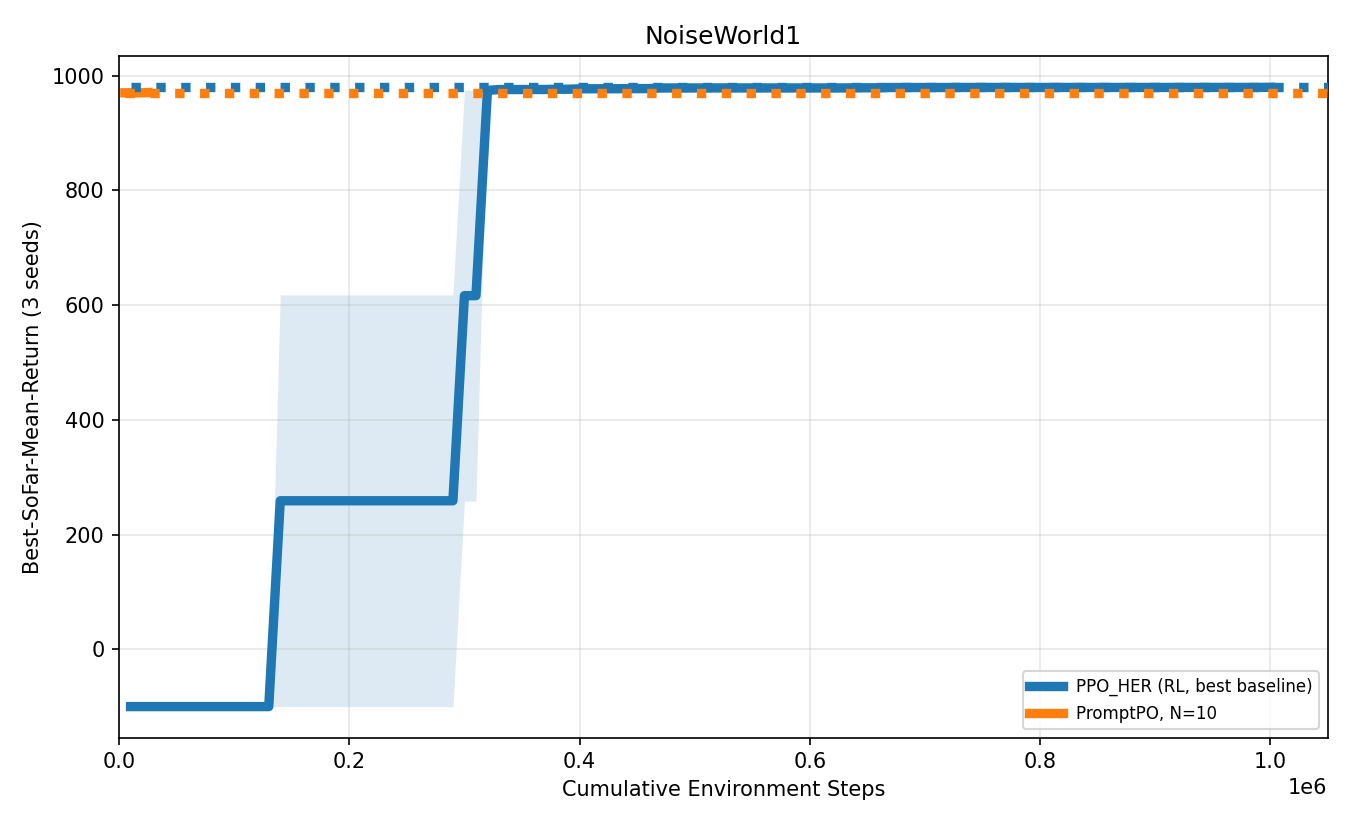}
        \caption{NoiseWorld1}
    \end{subfigure}
    \hfill
    \begin{subfigure}{0.19\textwidth}
        \centering
        \includegraphics[width=\linewidth]{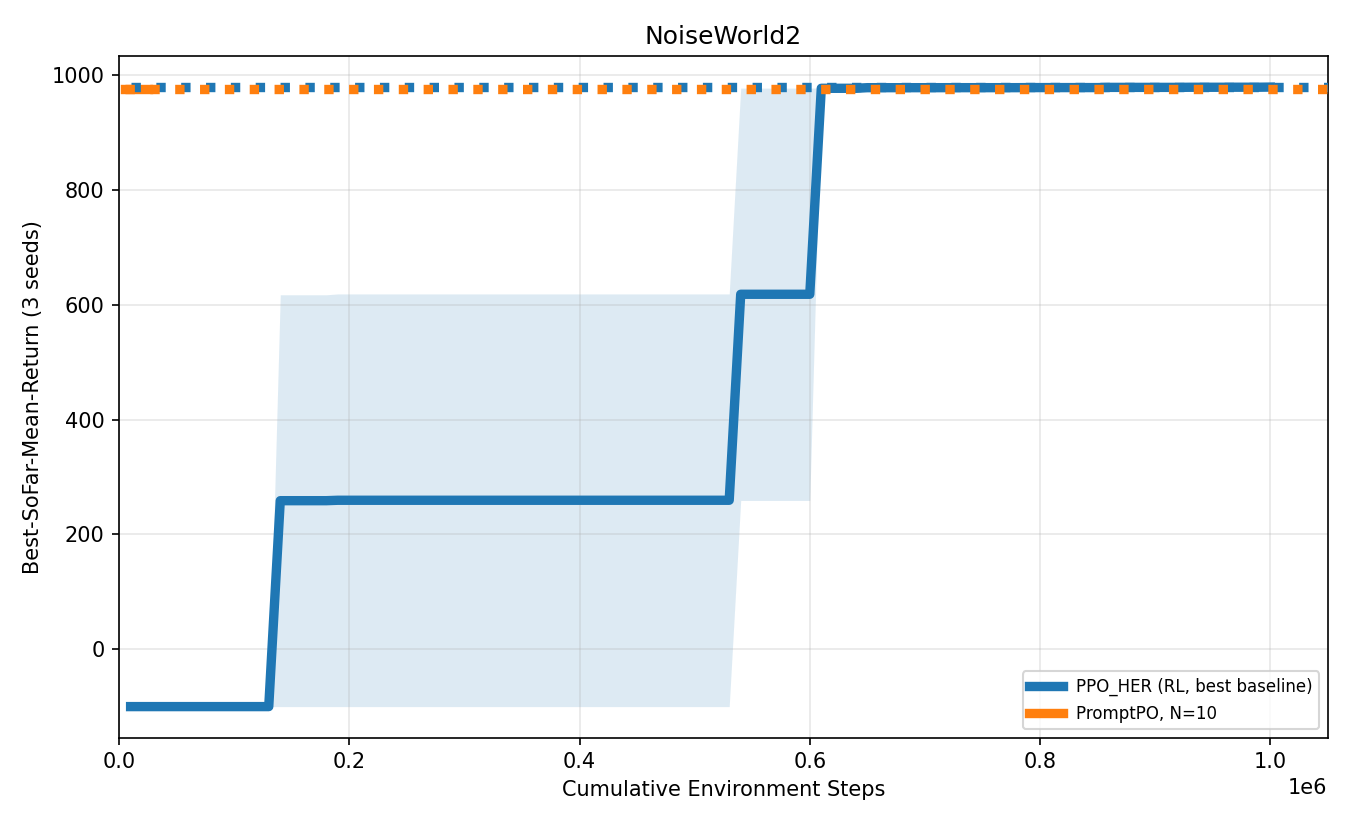}
        \caption{NoiseWorld2}
    \end{subfigure}
    \hfill
    \begin{subfigure}{0.19\textwidth}
        \centering
        \includegraphics[width=\linewidth]{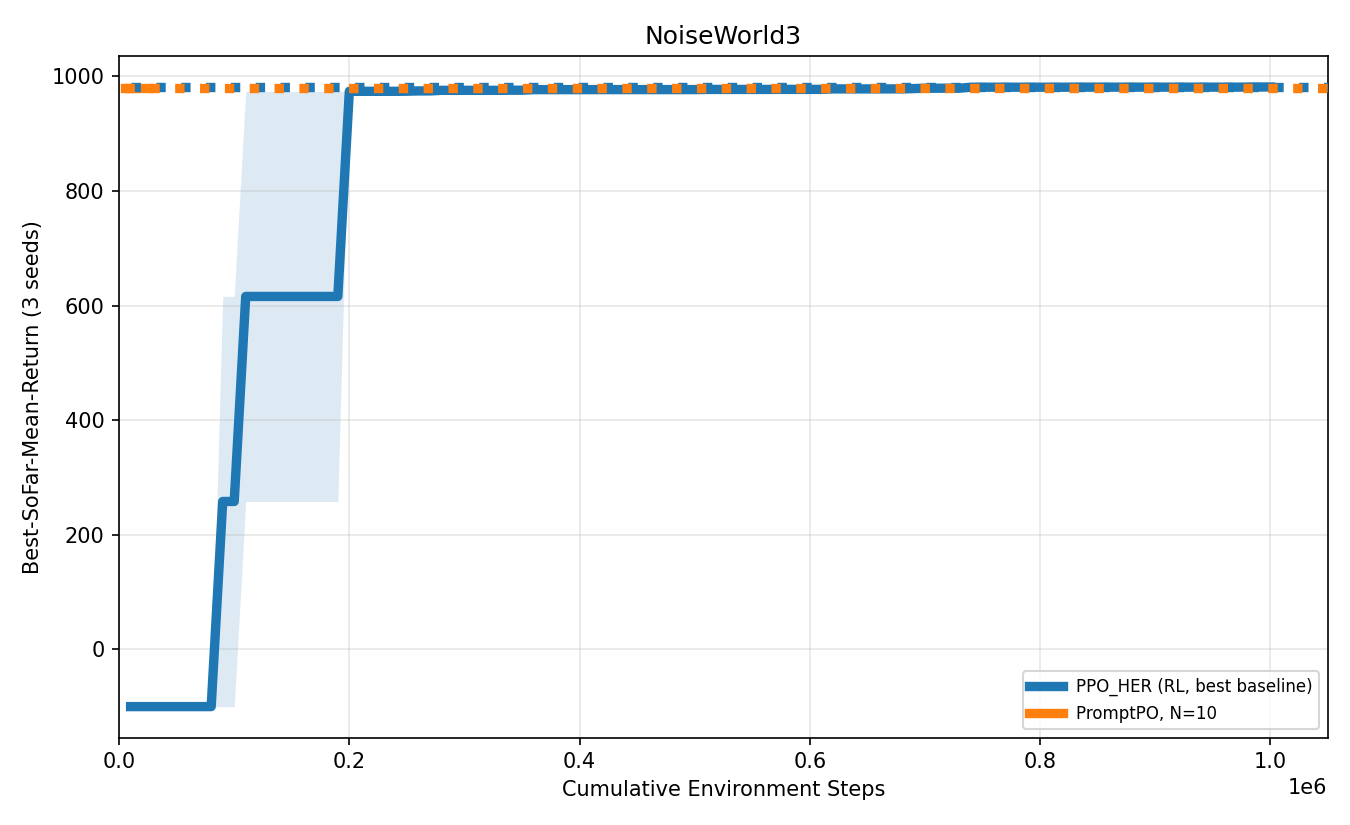}
        \caption{NoiseWorld3}
    \end{subfigure}
    \hfill
    \begin{subfigure}{0.19\textwidth}
        \centering
        \includegraphics[width=\linewidth]{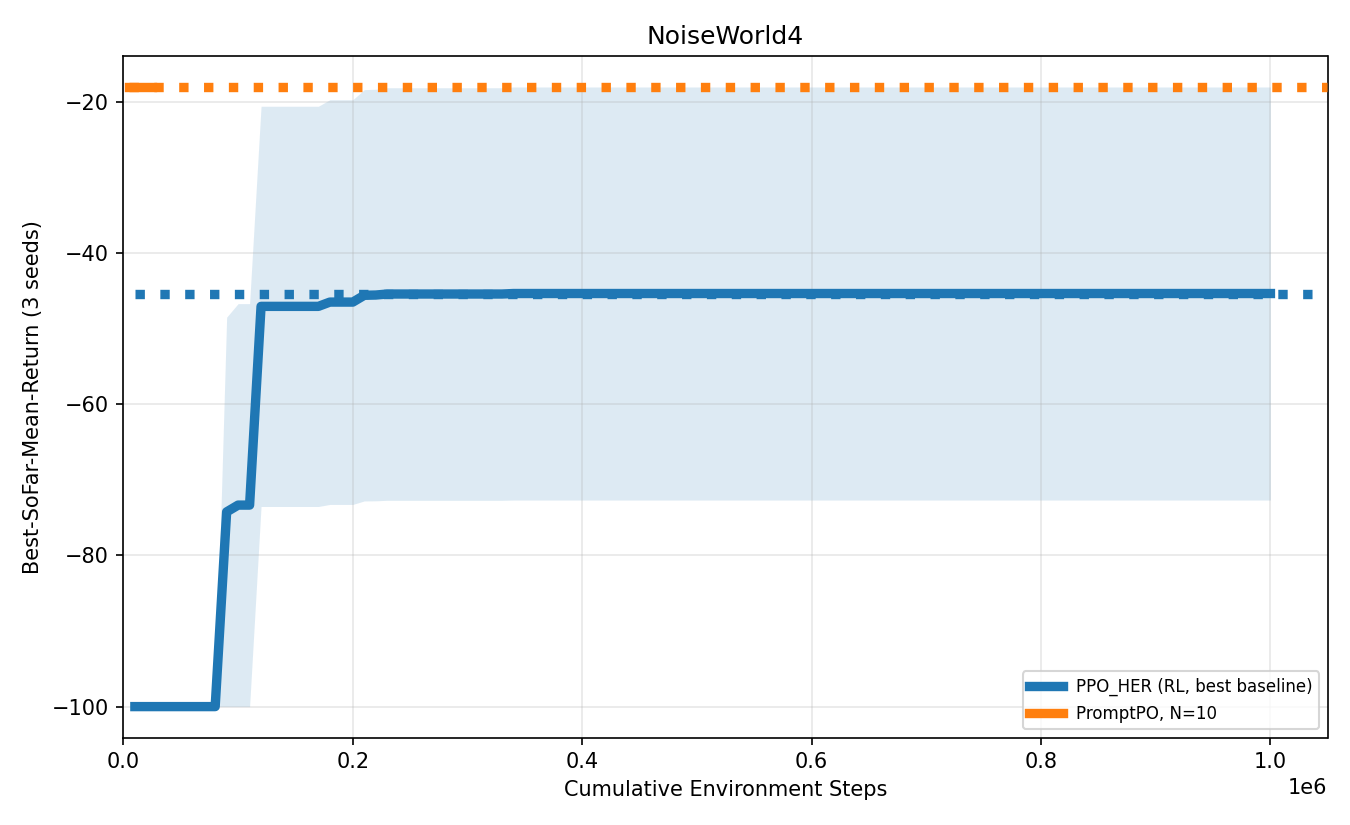}
        \caption{NoiseWorld4}
    \end{subfigure}
    \hfill
    \begin{subfigure}{0.19\textwidth}
        \centering
        \includegraphics[width=\linewidth]{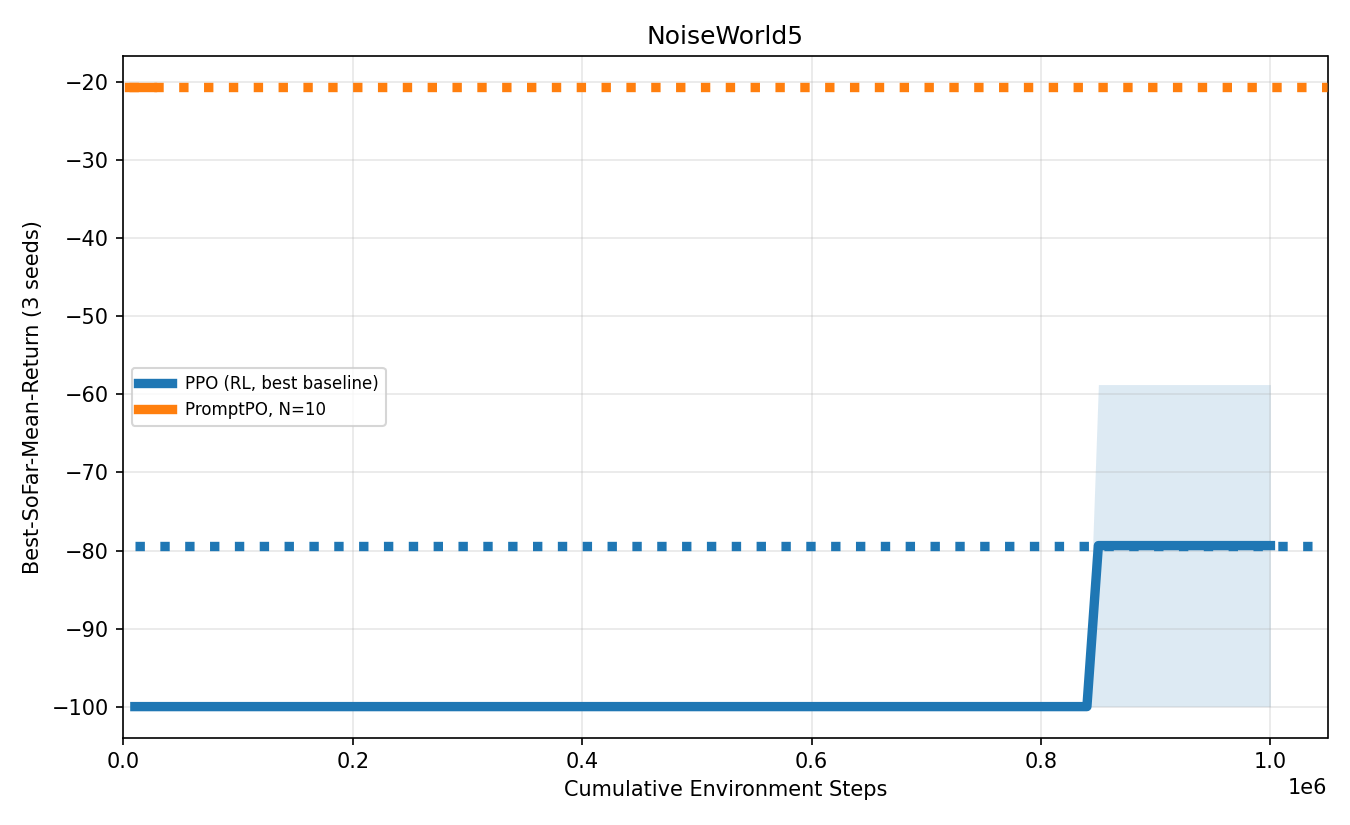}
        \caption{NoiseWorld5}
    \end{subfigure}
    \caption{Training curves across NoiseWorld boards for PromptPO and the best performing RL algorithm out of the set of methods we consider. Mean return is reported over 3 seeds. The dotted lines show best achieved final performance. In the rightmost two plots, unlike in the leftmost 3 plots, PromptPO and RL do not attain near-optimal performance.}
    \label{fig:noiseworldrow}
\end{figure}

\begin{wrapfigure}{r}{0.45\linewidth}
    \centering
    \includegraphics[width=\linewidth]{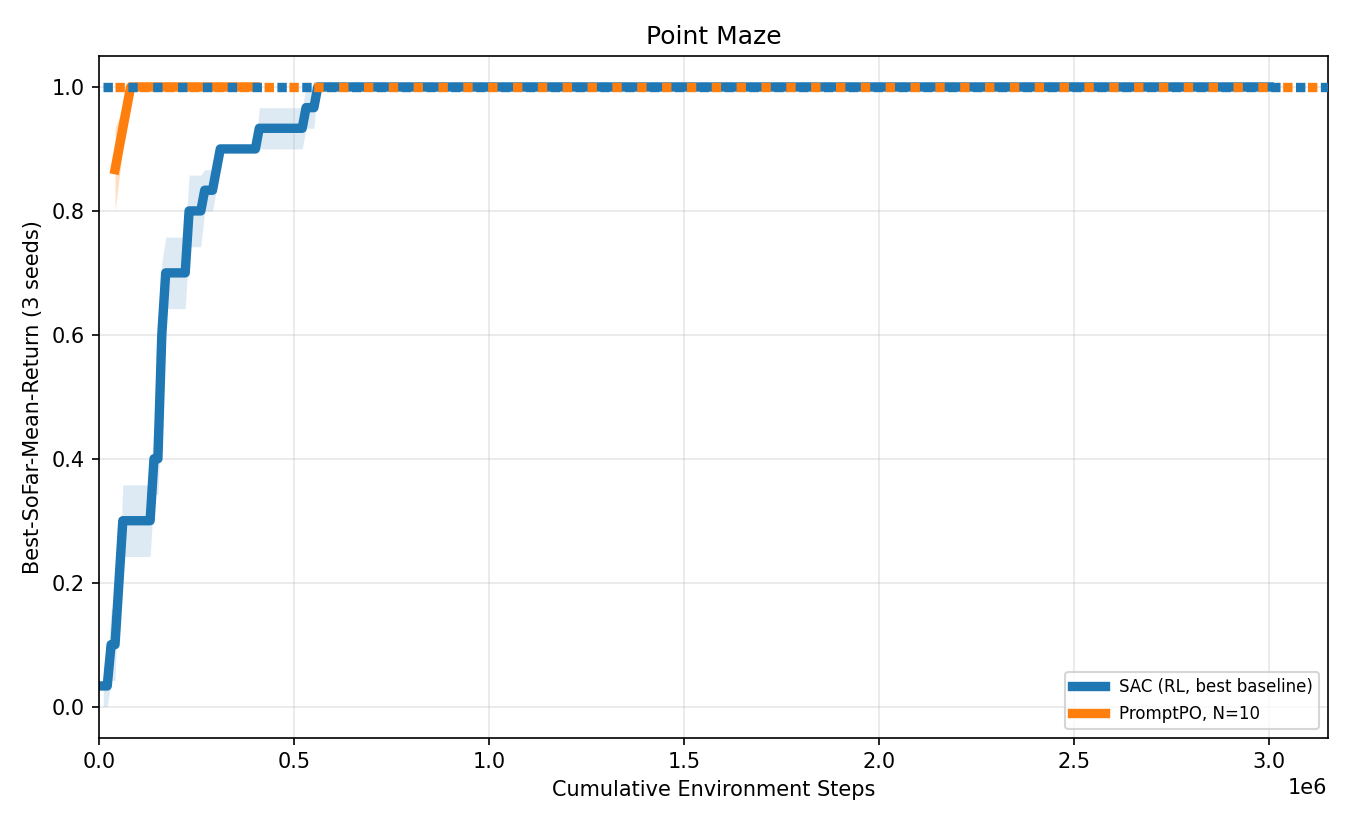}
    \caption{PromptPO's training performance in Point Maze versus SAC, which is the best performing RL algorithm out of the set of methods we consider. Mean return is reported over 3 seeds. The dotted lines show best achieved final performance.}
    \label{fig:pointmaze_training_curve}
\end{wrapfigure}

\textbf{Results~~} PromptPO matches the best performing RL algorithms performance with significantly fewer environment samples in all NoiseWorld environments and in Point Maze; in the NoiseWorld environment, PromptPO attains its best performing policy without updating its generated policies using feedback from the environment, despite no information about the different cell types. We attribute this strong performance to PromptPO's ability to sample many different policies, which in effect results in strong exploration capabilities. When the many different policies sampled by PromptPO don't immediately yield the best performing policy---as is the case in Point Maze---PromptPO incorporates the environment feedback (e.g., expressed via the return and generated evaluation function) to sample improved policies in the next round. Figure \ref{fig:pointmaze_training_curve} shows this learning capability for Point Maze, and Figure \ref{fig:noiseworldrow} show it for NoiseWorld.

Unlike in NoiseWorld1, NoiseWorld2, and NoiseWorld3, for NoiseWorld4 and NoiseWorld5, PromptPO and the best performing RL algorithm fail to find a policy that behaves near-optimally; learning to visit cell types 5 and 6 to unlock the high reward upon entering the goal state. This result indcates that PromptPO, like standard RL methods, struggle with difficult exploration problems; future work should try to incorporate lessons from exploration for RL into PromptPO. Providing PromptPO with context about cell types 5 and 6, i.e., that visiting them is necessary to attain the high reward bonus at the goal state, results in PromptPO producing a policy that exhibits this desired behavior. This type of prior knowledge might be difficult to specify to a standard RL algorithm, but is quite simple to specify to PromptPO via natural language and, as shown in Figure \ref{fig:noiseworld5abalation}, enables PromptPO to generate a performant policy.

\subsection{Robotics Environments}
\label{sec:robotics_envs}
\textbf{Mujoco Environments} We evaluate PromptPO in the Hopper, Ant, Half-Cheetah, and Swimmer environments, where the objective is to train a locomotion policy, and in the Reacher and Inverted Pendulum environments where the objective is to move a two-jointed robot arm close to a target and balance a two-jointed pole on a cart, respectively \citep{todorov2012mujoco}.

\textbf{Metaworld Environments} We evaluate PromptPO in the Button Press, Door Open, Drawer Open,  and Pick and Place environments \citep{yu2020metaworld}, where the objective is to train a 7-DOF Sawyer robotic arm to press a button, open a door, open a drawer, and pick and place objects, respectively.

For the Mujoco and Metaworld environments, we use the default observation and action spaces, and for PromptPO, additionally provide a description of what each observation variable denotes.


For Metaworld, we train policy with SAC, following the best performing method of \citet{yu2020metaworld}.

\begin{figure}[t]
    \centering
    \begin{subfigure}{0.24\textwidth}
        \centering
        \includegraphics[width=\linewidth]{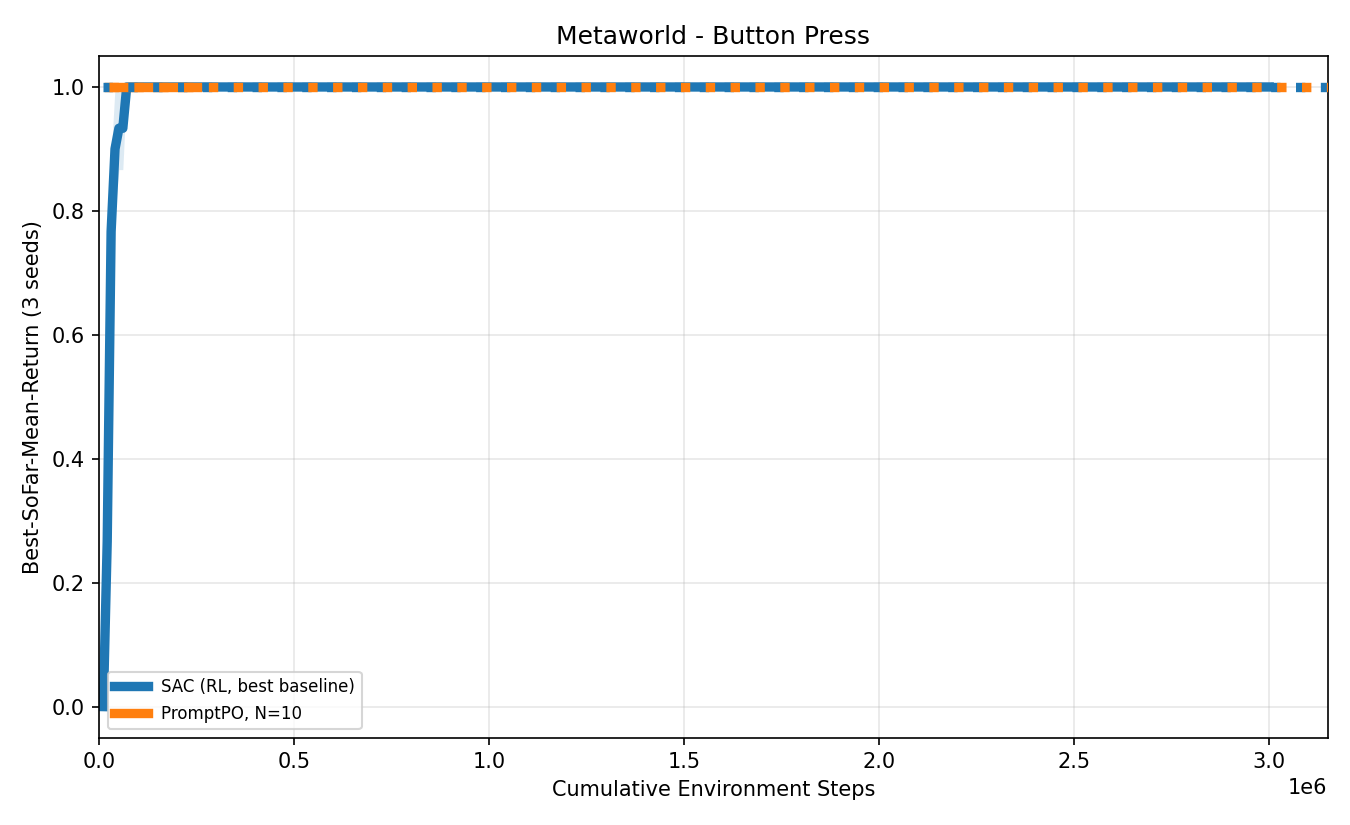}
        \caption{Button Press}
    \end{subfigure}
    \hfill
    \begin{subfigure}{0.24\textwidth}
        \centering
        \includegraphics[width=\linewidth]{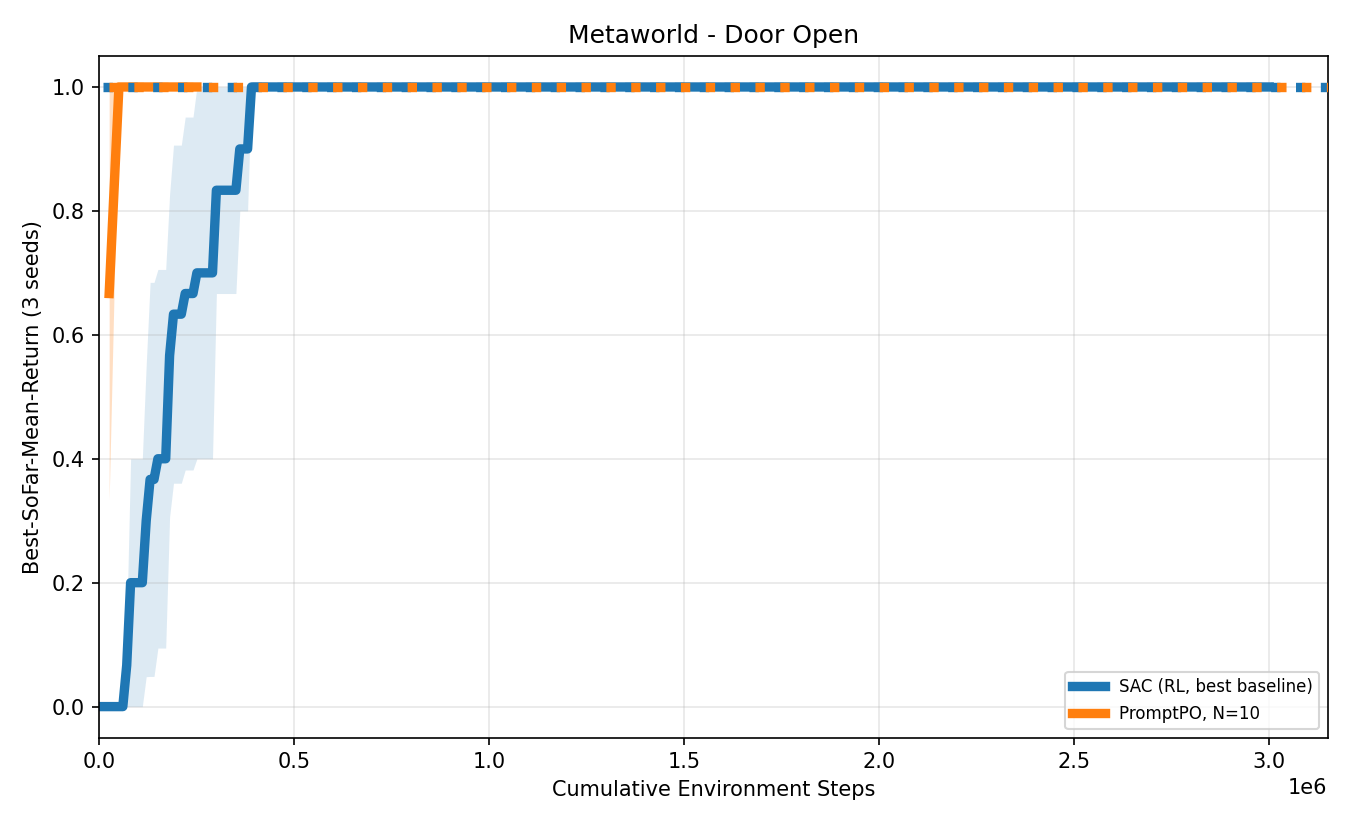}
        \caption{Door Open}
    \end{subfigure}
    \hfill
    \begin{subfigure}{0.24\textwidth}
        \centering
        \includegraphics[width=\linewidth]{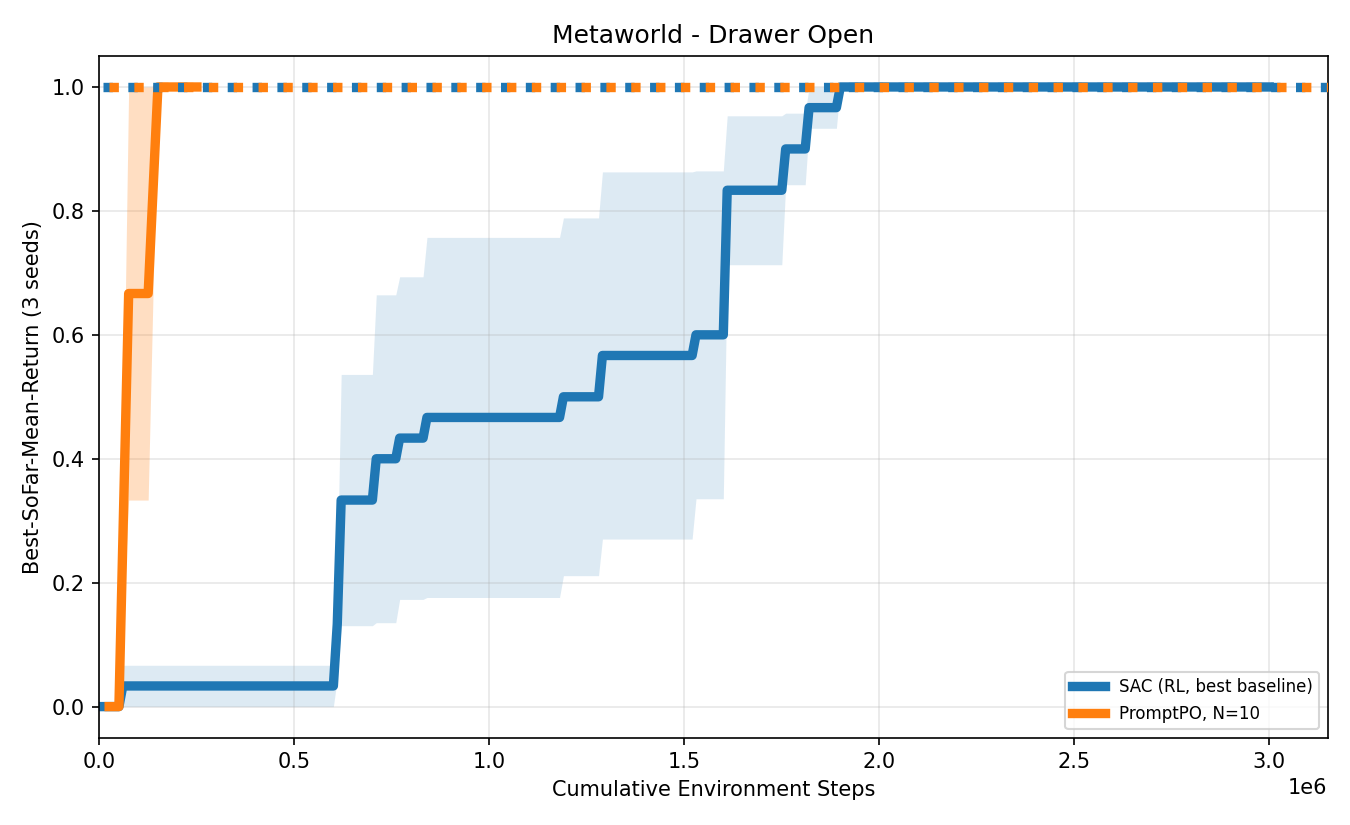}
        \caption{Drawer Open}
    \end{subfigure}
    \hfill
    \begin{subfigure}{0.24\textwidth}
        \centering
        \includegraphics[width=\linewidth]{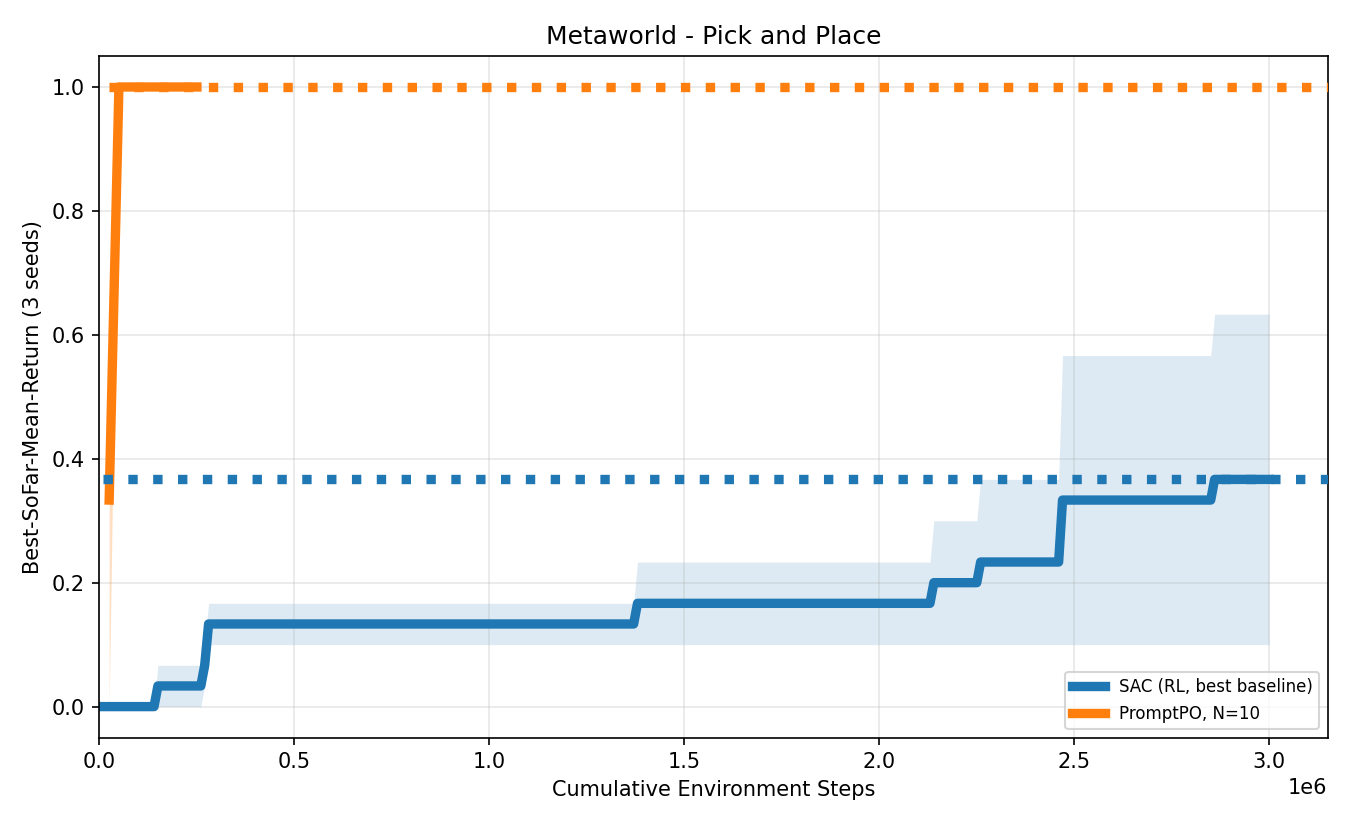}
        \caption{Pick Place}
    \end{subfigure}
    \caption{Training curves across Meta-World tasks for PromptPO and the best performing RL algorithm out of the set of methods we consider. Mean return is reported over 3 seeds. The dotted lines show best achieved final performance.}
    \label{fig:metaworld_row}
\end{figure}

\textbf{Results~~} PromptPO under performs RL in all Mujoco environments except Reacher, where it is substantially more sample efficient, and Swimmer, where it attains similar performance but after many more environment interactions. On the other hand, PromptPO is significantly more sample efficient than RL in all Metaworld environments, and additionally outperforms the final performance attained by RL in the Pick and Place environment. In Metaworld, PromptPO generates its best performing policy after incorperating only a single or a few rounds of feedback from the environment, as illustrated in Figure \ref{fig:metaworld_row}.

\begin{figure}[t]
    \centering
    \begin{subfigure}{0.32\textwidth}
        \centering
        \includegraphics[width=\linewidth]{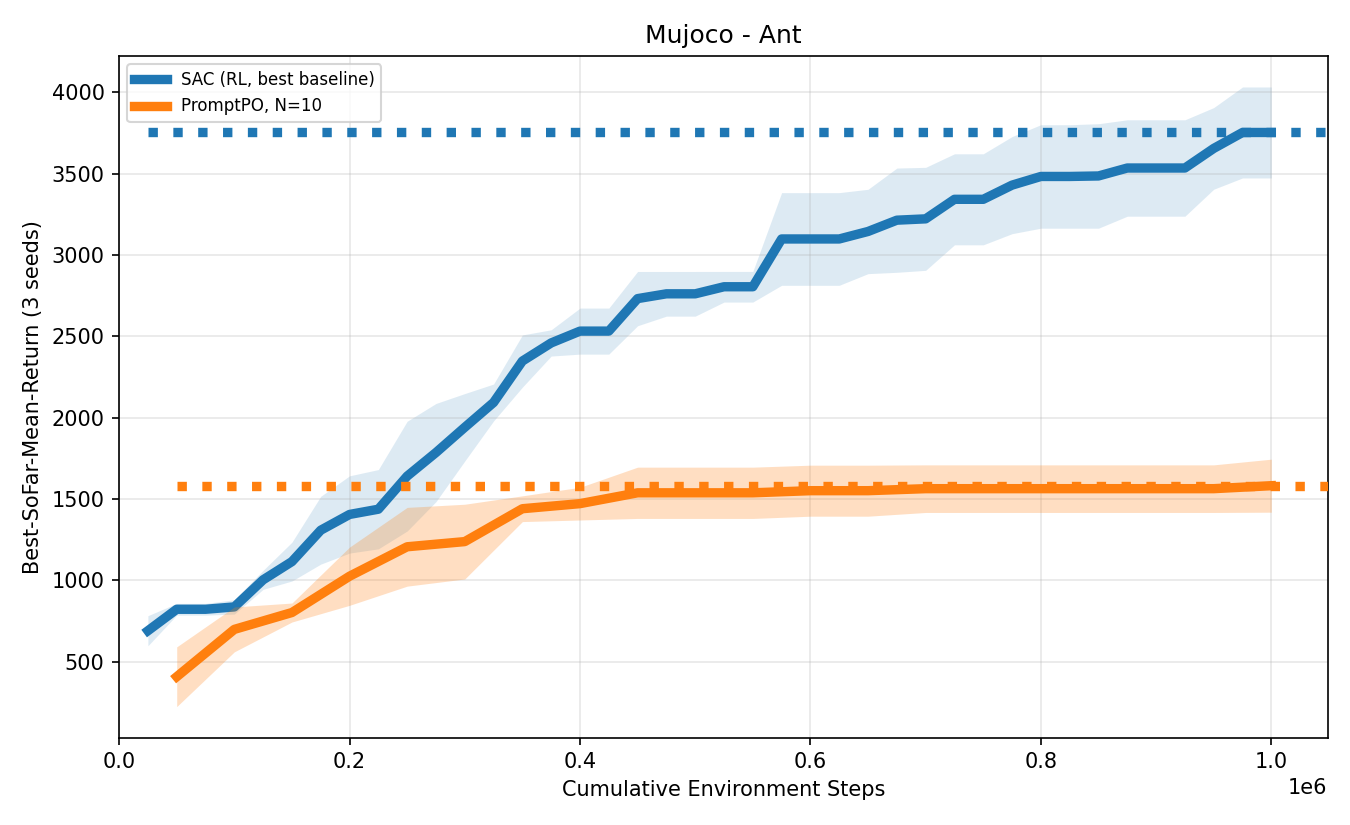}
        \caption{Ant}
    \end{subfigure}
    \hfill
    \begin{subfigure}{0.32\textwidth}
        \centering
        \includegraphics[width=\linewidth]{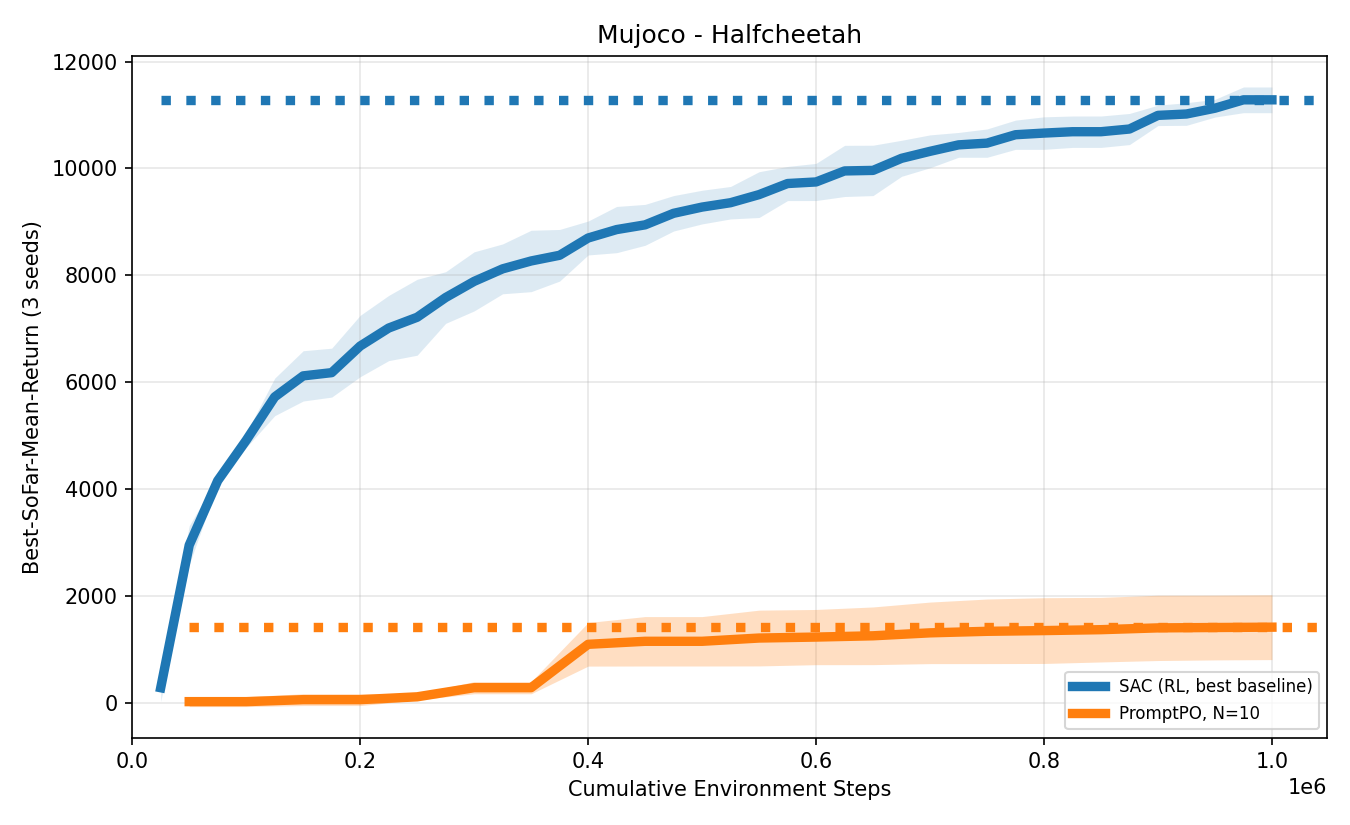}
        \caption{HalfCheetah}
    \end{subfigure}
    \hfill
    \begin{subfigure}{0.32\textwidth}
        \centering
        \includegraphics[width=\linewidth]{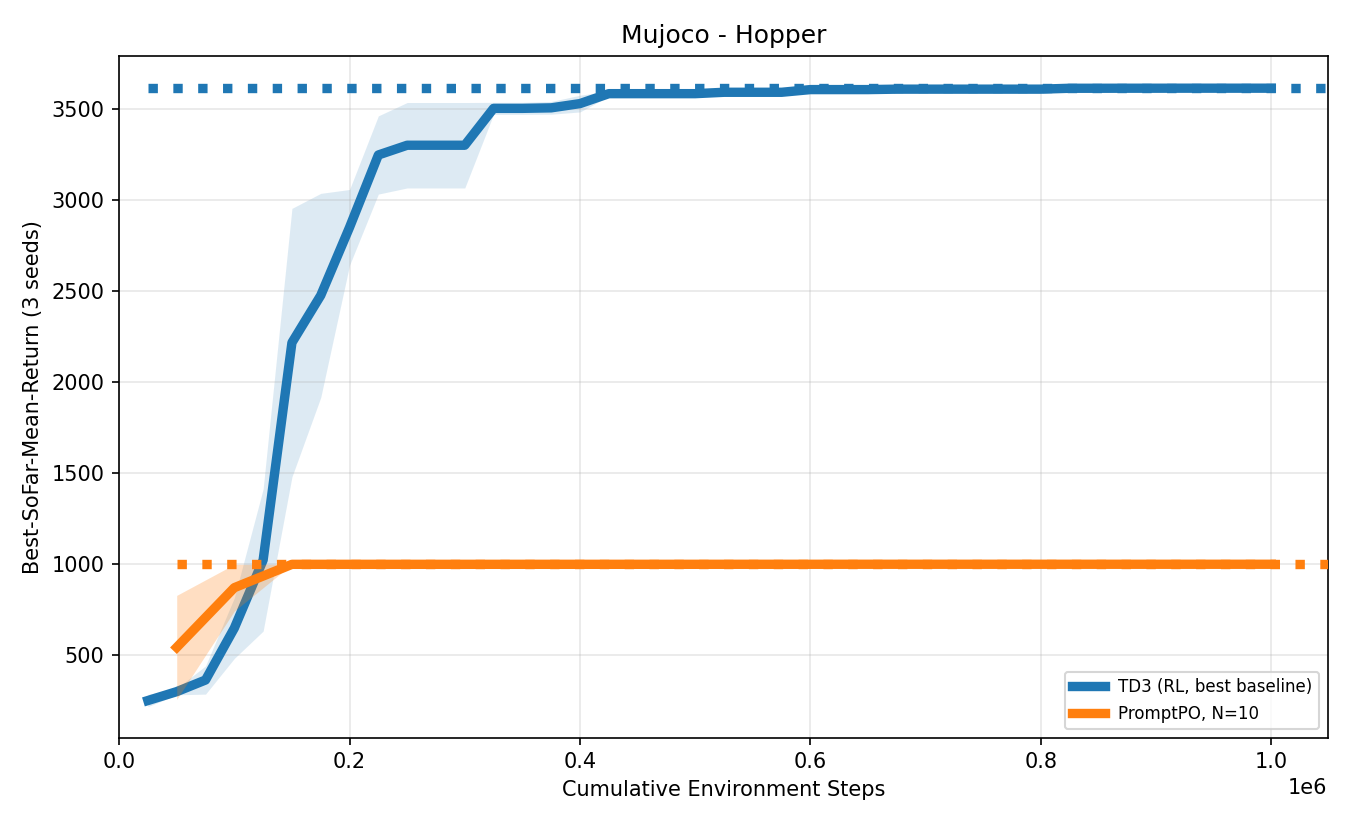}
        \caption{Hopper}
    \end{subfigure}

    \vspace{0.5em}

    \begin{subfigure}{0.32\textwidth}
        \centering
        \includegraphics[width=\linewidth]{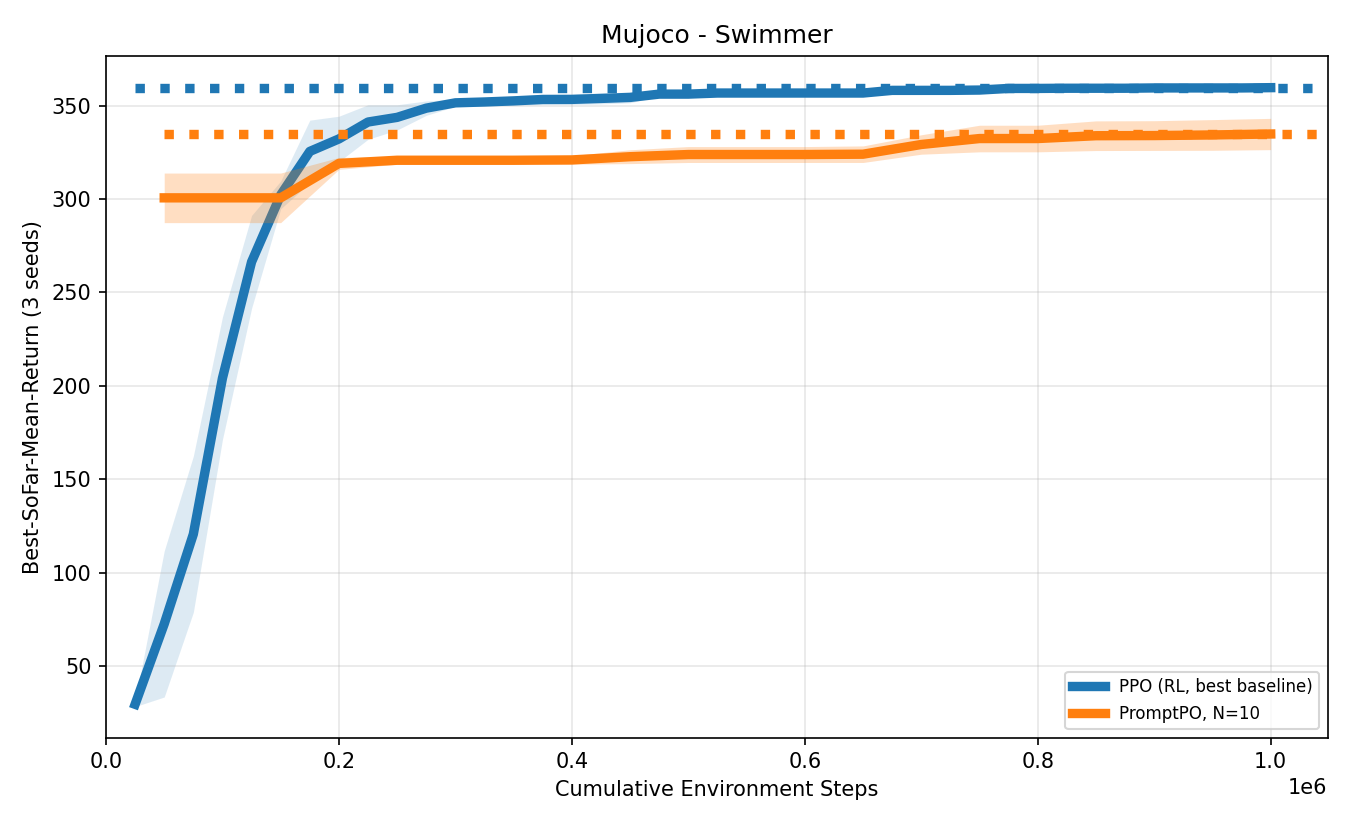}
        \caption{Swimmer}
    \end{subfigure}
    \hfill
    \begin{subfigure}{0.32\textwidth}
        \centering
        \includegraphics[width=\linewidth]{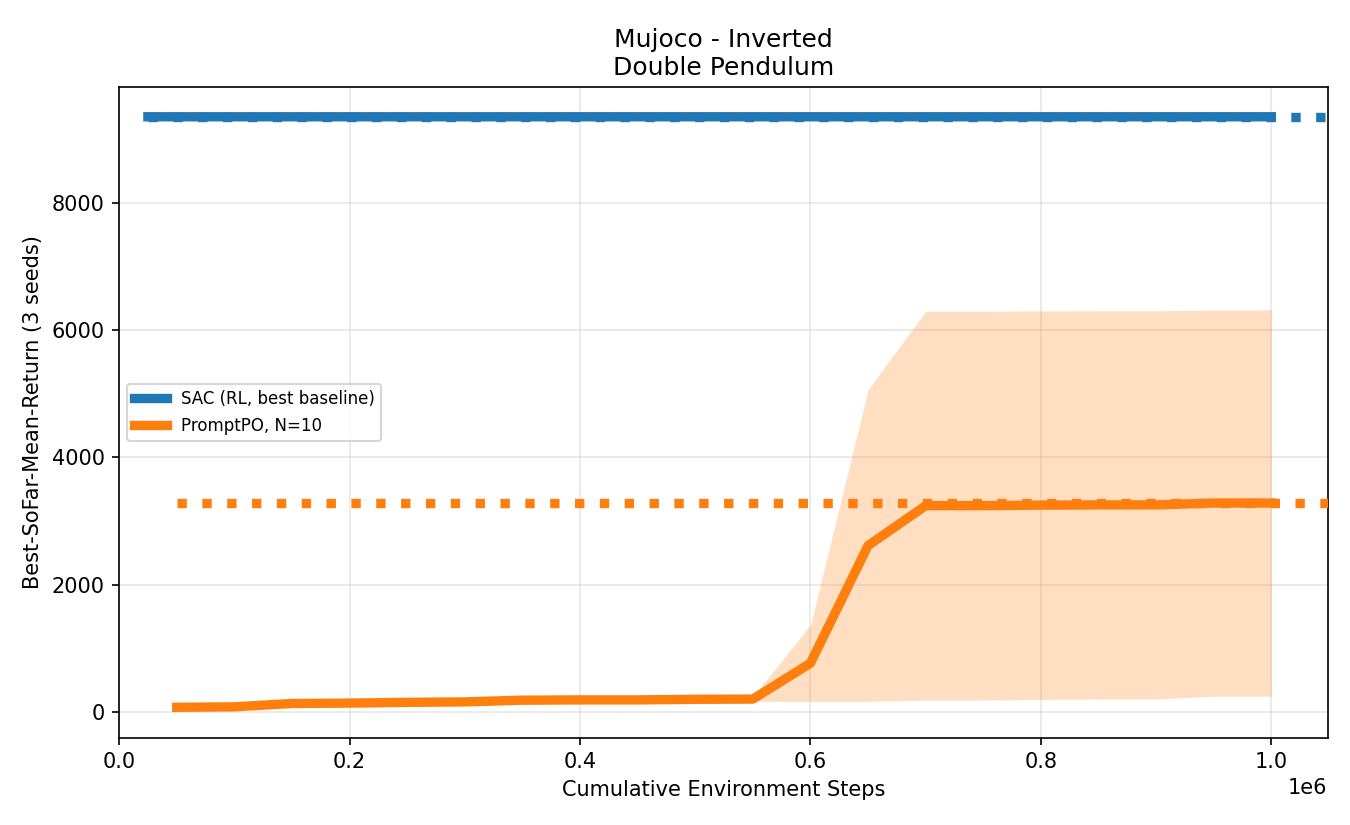}
        \caption{Inverted Double Pendulum}
    \end{subfigure}
    \hfill
    \begin{subfigure}{0.32\textwidth}
        \centering
        \includegraphics[width=\linewidth]{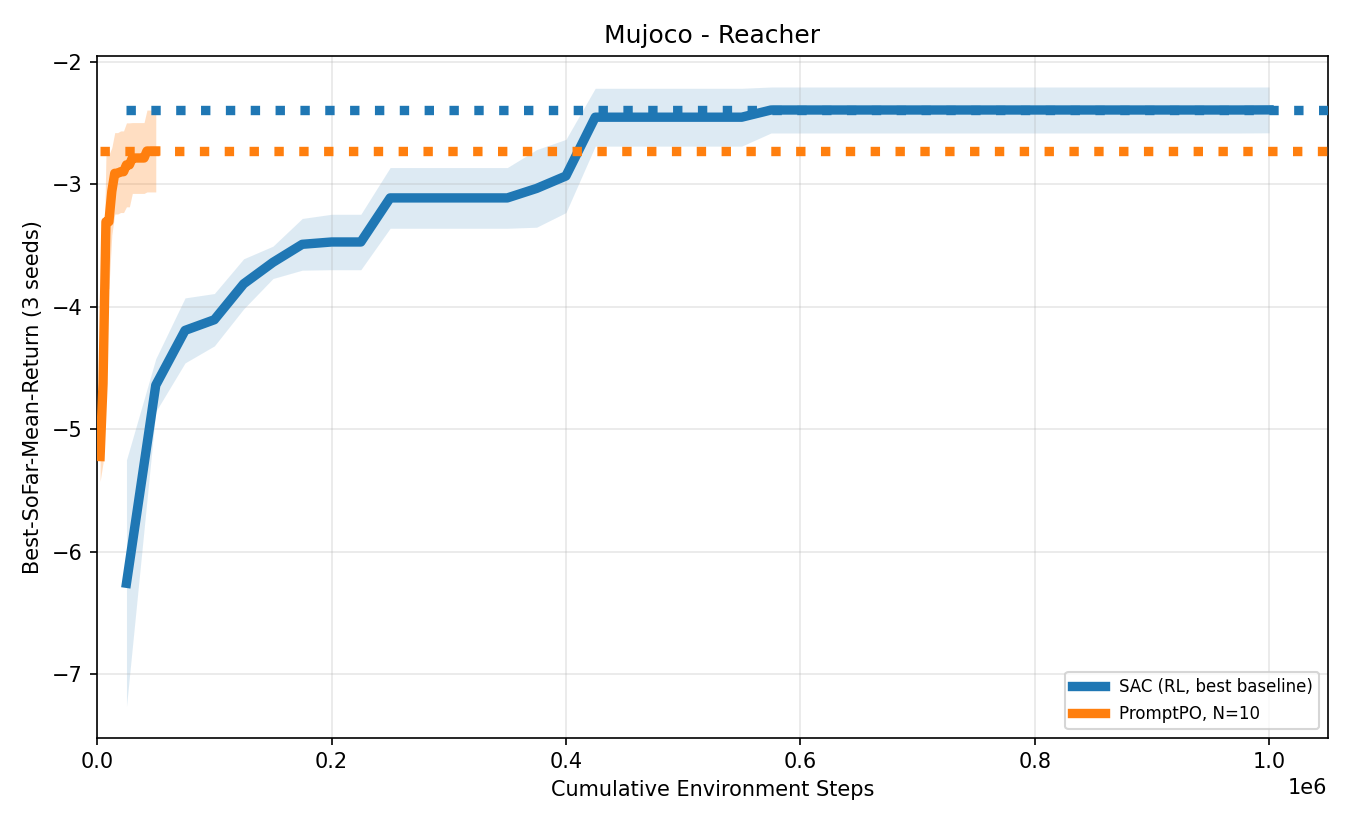}
        \caption{Reacher}
    \end{subfigure}
    \caption{Training curves across MuJoCo continuous control tasks for PromptPO and the best performing RL algorithm out of the set of methods we consider. Mean return is reported over 3 seeds. The dotted lines show best achieved final performance.}
    \label{fig:mujoco_training}
\end{figure}

The difference in relative performance between PromptPO and RL in the Mujoco and Metaworld environments provide insight into the type of environments where PromptPO is a sufficient policy optimizer; for Mujoco, the action space consists of torques applied to hinge joints, while for Metaworld it is the end-effector displacement and gripper finger positions. We hypothesize that PromptPO struggles to implement a performant policy for environments that require fine-grained continuous control, and hence it performs poorly in Mujoco environments while exhibiting strong performance in Metaworld environments. For all environments, PromptPO implements a proportional controller and tunes the controller after subsequent environment feedback is provided in context. This suggests that, for some tasks, PromptPO restricts its policy class.

\subsection{Real World Control Tasks}
\label{sec:real_world_envs}
\textbf{Glucose Monitoring~~} In Glucose Monitoring, the objective is to design a policy to administer insulin to a patient with Type II diabetes; an action must be outputted every 5 minutes for 20 days, and observations are provided by a continuous glucose monitor (CGM). We use the environment parameters and reward function from \citet{pan2022effects}.

\textbf{Pandemic Mitigation} In Pandemic Mitigation, the objective is to design COVID-19 pandemic lockdown
regulations to balance economic and health outcomes. The available actions are four lockdown regulation stages, and the observations consist of the states from an SEIRS pandemic model. We use the Pandemic simulator from \citet{kompella2020reinforcement} and the  environment parameters and reward function from \citet{pan2022effects}.

\textbf{Traffic Control} In Traffic Control, a policy must control a fleet of autonomous vehicles merging onto a highway to maximize traffic flow. The action space consists of the autonomous vehicle accelerations, the observation space is location and velocity---as well as other statistics derived from these measures---of all cars on the highway. We use the simulator from \citet{9489303} and the  environment parameters and reward function from \citet{pan2022effects}.

\begin{figure}[t]
    \centering
    \begin{subfigure}{0.32\textwidth}
        \centering
        \includegraphics[width=\linewidth]{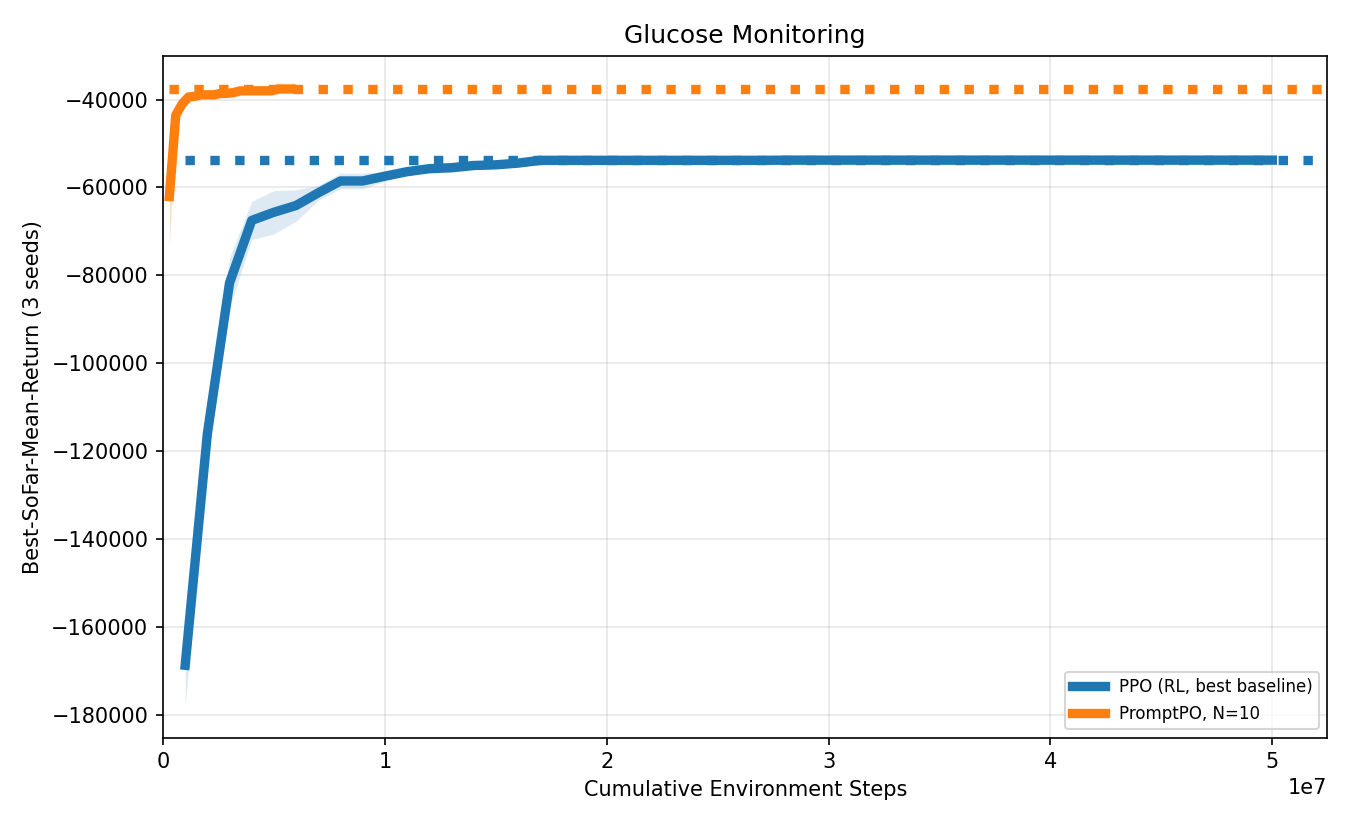}
        \caption{Glucose Monitoring}
    \end{subfigure}
    \hfill
    \begin{subfigure}{0.32\textwidth}
        \centering
        \includegraphics[width=\linewidth]{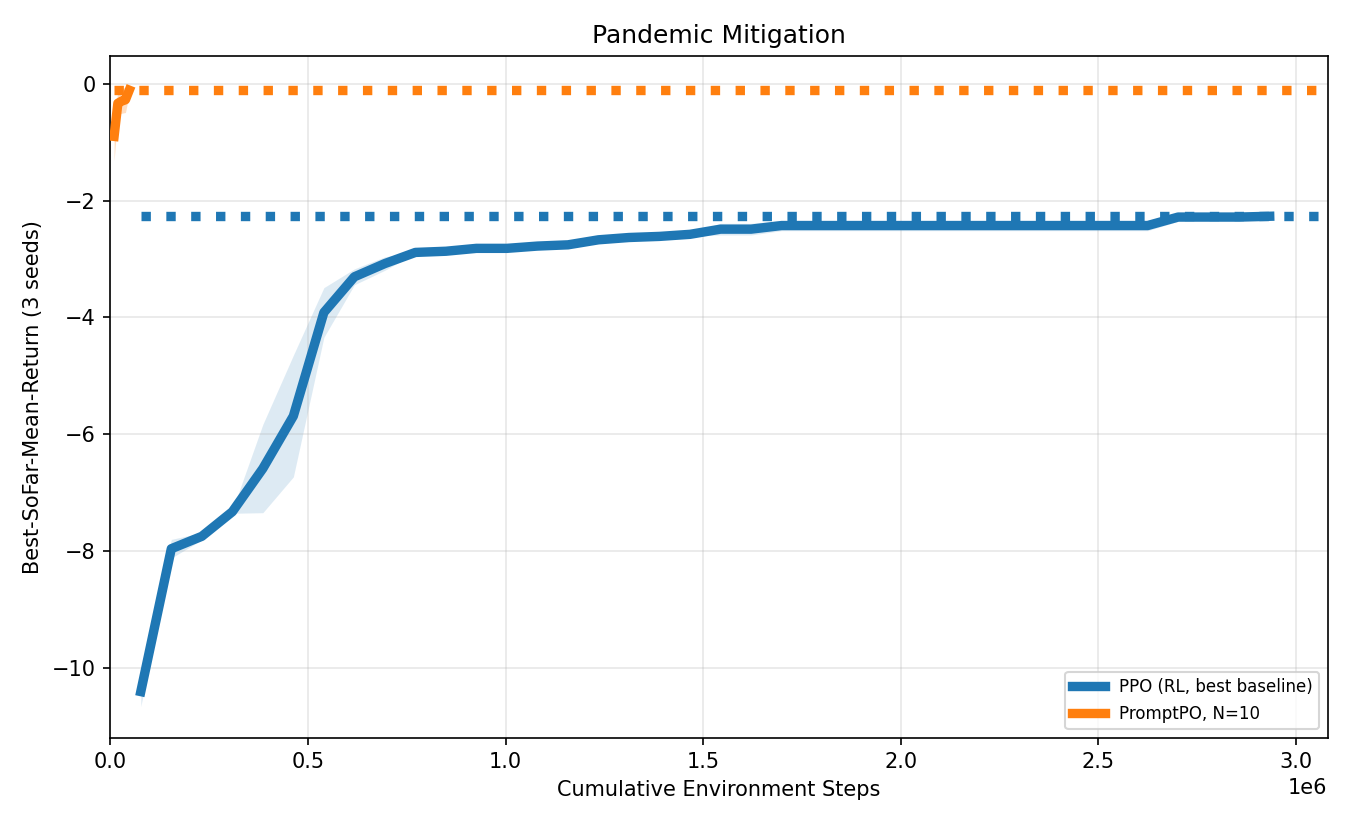}
        \caption{Pandemic Mitigation}
    \end{subfigure}
    \hfill
    \begin{subfigure}{0.32\textwidth}
        \centering
        \includegraphics[width=\linewidth]{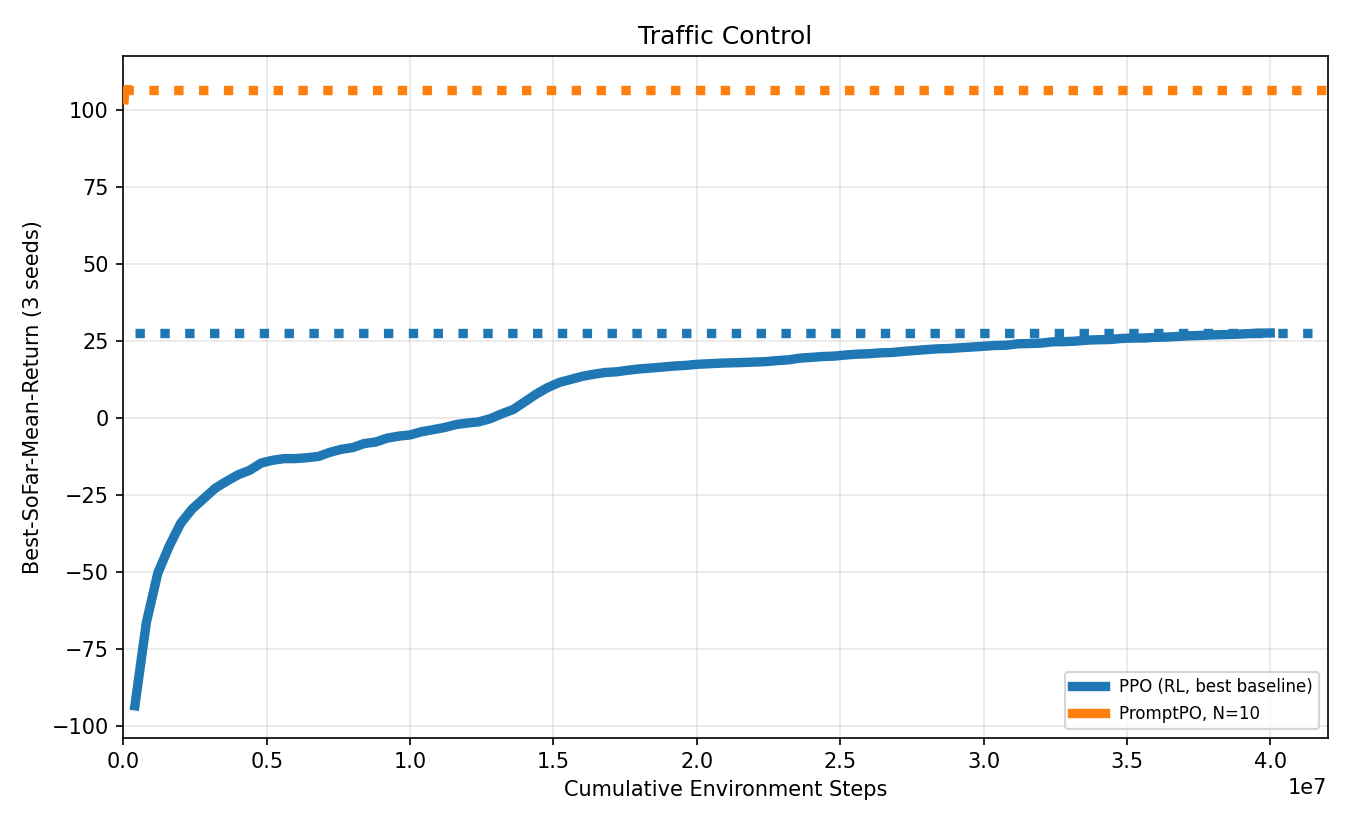}
        \caption{Traffic Control}
    \end{subfigure}
    \caption{Training curves across real-world control environments for PromptPO and PPO. Mean return is reported over 3 seeds. The dotted lines show best achieved final performance.}
    \label{fig:real_world_row}
\end{figure}

\textbf{Results} For these environments, representative of many real world control tasks, PromptPO is both significantly more sample efficient than RL and, for Traffic and Glucose environments, substantially outperforms the best policy found by RL within the environment sample budget considered. Training curves for all methods are shown in Figure \ref{fig:real_world_row}. These results showcase PromptPO's strength as a policy optimizer in real world settings where it may be capable of leveraging its pretraining data as a strong prior on generating a performant policy.

\section{Prompts used by PromptPO}

\begin{figure}[H]
\centering
\begin{tcolorbox}[
    colback=gray!5,
    colframe=gray!50,
    boxrule=0.4pt,
    arc=2pt,
    left=4pt,
    right=4pt,
    top=4pt,
    bottom=4pt,
    title={Trajectory Feedback Prompt},
    fonttitle=\bfseries,
    width=0.95\linewidth
]
\footnotesize
\ttfamily
Implement a class called Feedback with a method summarize\_trajectory(self, traj).\\[0.4em]
traj is a list of observation objects (each a \{obs\_cls\} instance) for one episode in time order.\\[0.4em]
Compute brief statistics to help improve a policy maximizing expected reward under \{reward\_function\_name\} over \{n\_timesteps\} timesteps.\\[0.4em]
Examples: sum, min, max, mean of reward-related quantities or other useful summaries.\\[0.4em]
\end{tcolorbox}
\caption{Prompt used to generate trajectory-level feedback summaries for improving policies.}
\label{fig:feedback_prompt}
\end{figure}

\begin{figure}[H]
\centering
\begin{tcolorbox}[
    colback=gray!5,
    colframe=gray!50,
    boxrule=0.4pt,
    arc=2pt,
    left=4pt,
    right=4pt,
    top=4pt,
    bottom=4pt,
    title={Policy Generation Prompt},
    fonttitle=\bfseries,
    width=0.97\linewidth
]
\footnotesize
\ttfamily
Observation Context: \{obs\_context\}\\[0.4em]
Implementation Details: \{policy\_context\}\\[0.4em]
\{reward\_function\_name\}: \{reward\_src\}\\[0.6em]
\{history\_of\_previously\_generated\_best\_policies\}\\[0.6em]

Given the \{reward\_function\_name\}, implement a policy in python that inputs an observation and outputs an action. The policy should maximize the expected sum of rewards with respect to the \{reward\_function\_name\} over \{n\_timesteps\} timesteps. Think step-by-step internally before producing final code. Implement the policy in a class called \{policy\_class\_name\} with a function act(obs) that takes in the observation and returns a valid action.
\end{tcolorbox}
\caption{Prompt used to instruct the language model to generate a policy implementation from observation context, implementation details, and a reward function specification.}
\label{fig:policy_generation_prompt}
\end{figure}

\begin{figure}[H]
\centering
\begin{tcolorbox}[
    colback=gray!5,
    colframe=gray!50,
    boxrule=0.4pt,
    arc=2pt,
    left=4pt,
    right=4pt,
    top=4pt,
    bottom=4pt,
    title={Policy Evaluation Prompt},
    fonttitle=\bfseries,
    width=0.97\linewidth
]
\footnotesize
\ttfamily
\{reward\_function\_name\}: \{reward\_src\}\\[0.6em]

Latest policy to assess (attempt \{current\_round\_index\}) source:\\
\{generated\_policy\}\\[0.6em]
Latest policy (attempt \{current\_round\_index\}) episode returns (sum of rewards over each rollout): \{list(episode\_returns)\}, mean return \{mean\_return\_latest:.6f\}.\\[0.6em]

\{history\_of\_previously\_generated\_policies\}\\

Compare the latest policy's returns to the earlier attempts listed above. Reply in 3--4 sentences total. Your first sentence must open with a comparison: for example that this policy did better than certain prior attempts, worse than certain prior attempts, or performed similarly to them (name attempt numbers if helpful).

If the latest mean return is clearly very low for the \{horizon\}-timestep rollout, or clearly far below the best prior mean return in this run, or otherwise obviously suboptimal, open instead with that this policy did poorly (or equivalent), rather than implying improvement.

Then briefly relate this to maximizing the expected sum of rewards under the \{reward\_function\_name\} over \{horizon\} timesteps.

Reply with only that explanation, no code.
\end{tcolorbox}
\caption{Prompt used to elicit concise natural language evaluations of generated policies, comparing performance across attempts and relating outcomes to the target reward function.}
\label{fig:policy_evaluation_prompt}
\end{figure}

\section{Hyperparameter tuning method for NoiseWorld}
\label{app:noiseworld_hyperparams}
On each NoiseWorld board, we tune PPO as implemented in Stable-Baselines3 \citep{raffin2021stable}. For PPO we grid-search learning rate, PPO clip range, minibatch size, and optimization epochs (three values each; $3^4{=}81$ configurations per board). Every configuration is trained for a fixed environment step budget with periodic deterministic evaluation rollouts; we record the best mean evaluation return along each run and the environment step at which that peak first occurs. We repeat each configuration with three random seeds and, for a given board, select the hyperparameters that maximize the mean (across seeds) of those per-seed highest returns, breaking ties by lower mean timestep-to-highest return.

For distributional QR-DQN (Rainbow-style QRDQN) we use default hyperparameter from Stable-Baselines3 due to computational and time limitations.

\section{Varying the number of samples generated by PromptPO}
\label{app:varrying_n}
\begin{figure}[H]
    \centering
    \begin{subfigure}[t]{0.32\linewidth}
        \centering
        \includegraphics[width=\linewidth]{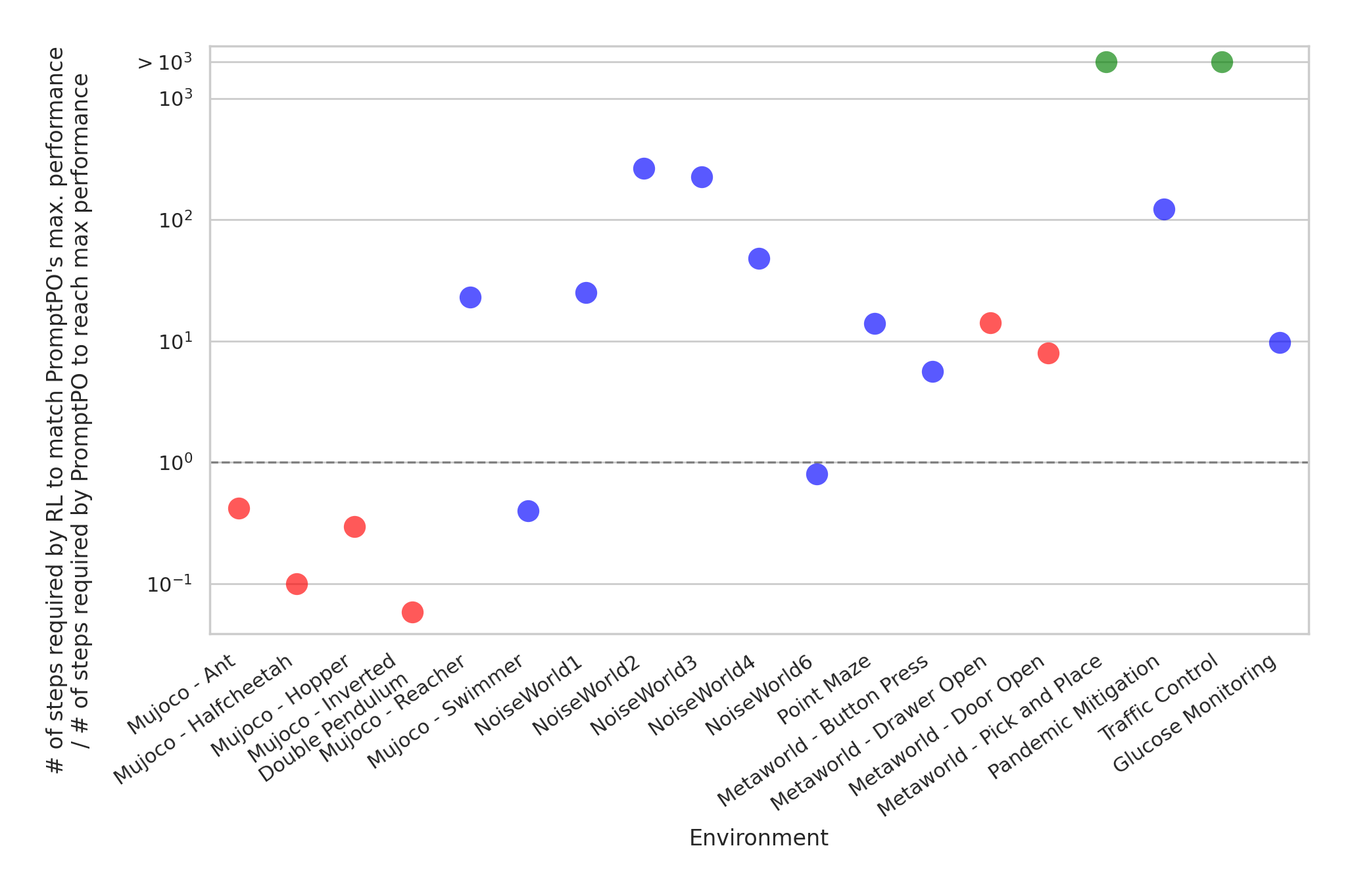}
        \caption{$N=5$}
        \label{fig:sample_efficiency_summary_n5}
    \end{subfigure}
    \hfill
    \begin{subfigure}[t]{0.32\linewidth}
        \centering
        \includegraphics[width=\linewidth]{images/plot2_sample_efficiency_n_gen_10.png}
        \caption{$N=10$}
        \label{fig:sample_efficiency_summary_n10}
    \end{subfigure}
    \hfill
    \begin{subfigure}[t]{0.32\linewidth}
        \centering
        \includegraphics[width=\linewidth]{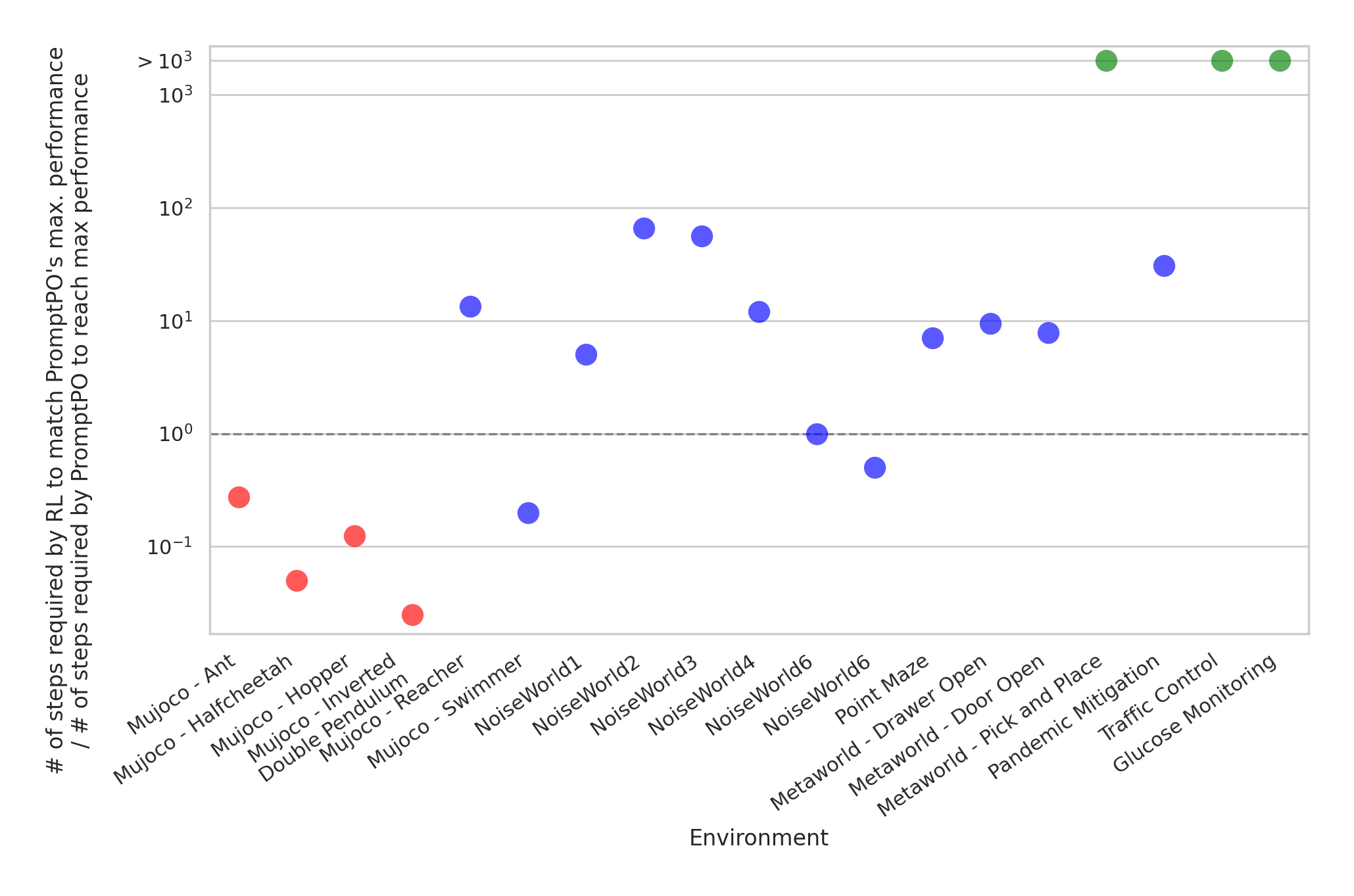}
        \caption{$N=20$}
        \label{fig:sample_efficiency_summary_n20}
    \end{subfigure}
    \caption{
    PromptPO performance summary for different numbers of sampled candidate policies per round ($N \in \{5,10,20\}$).
    This figure complements Figure~\ref{fig:sample_efficiency_summary}; see its caption for details on interpretation.
    }
    \label{fig:sample_efficiency_summary_combined}
\end{figure}

\begin{figure}[H]
    \centering
    \begin{subfigure}[t]{0.32\linewidth}
        \centering
        \includegraphics[width=\linewidth]{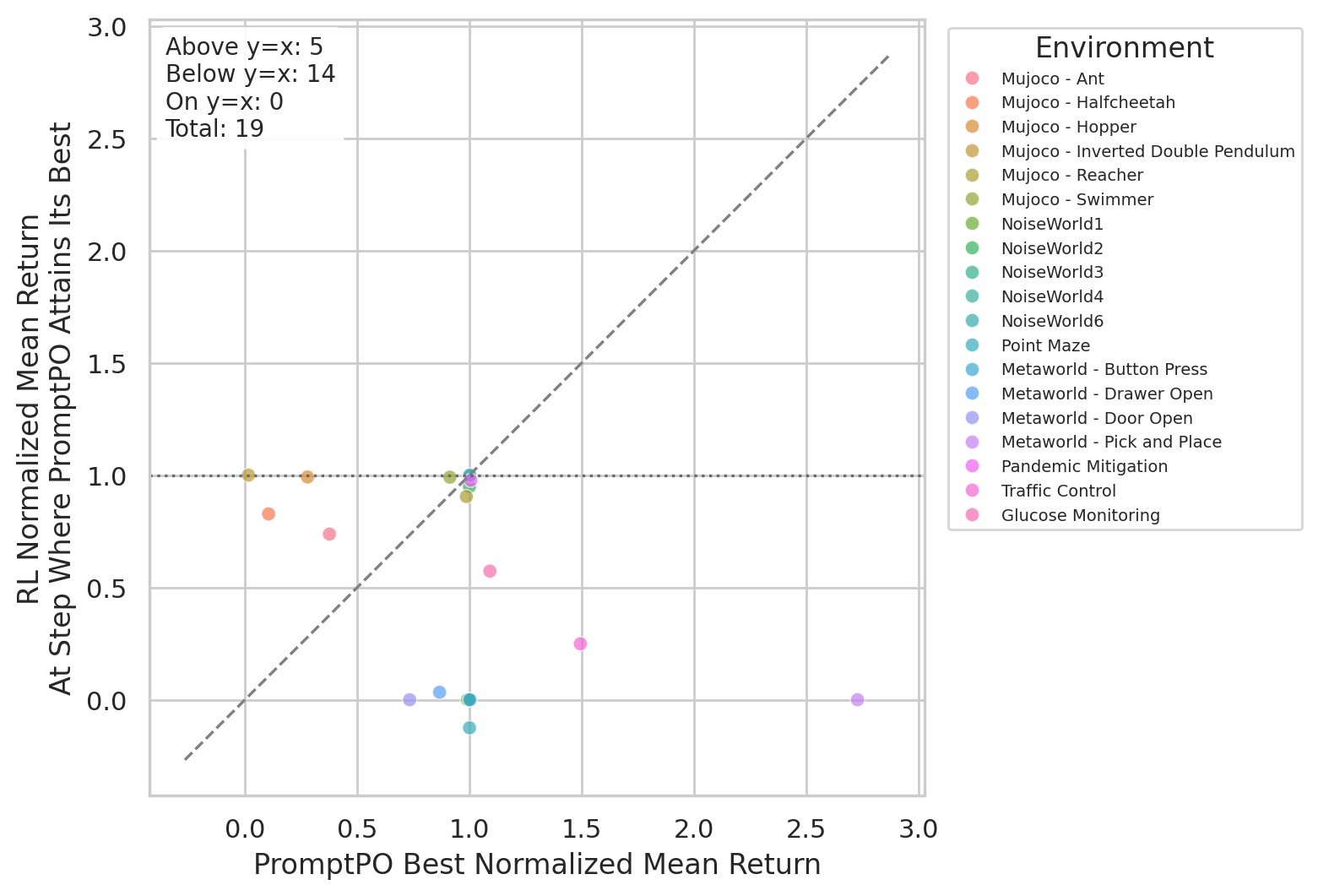}
        \caption{$N=5$}
        \label{fig:norm_mean_ret_summary_n5}
    \end{subfigure}
    \hfill
    \begin{subfigure}[t]{0.32\linewidth}
        \centering
        \includegraphics[width=\linewidth]{images/plot1_performance_scatter_n_gen_10.png}
        \caption{$N=10$}
        \label{fig:norm_mean_ret_summary_n10}
    \end{subfigure}
    \hfill
    \begin{subfigure}[t]{0.32\linewidth}
        \centering
        \includegraphics[width=\linewidth]{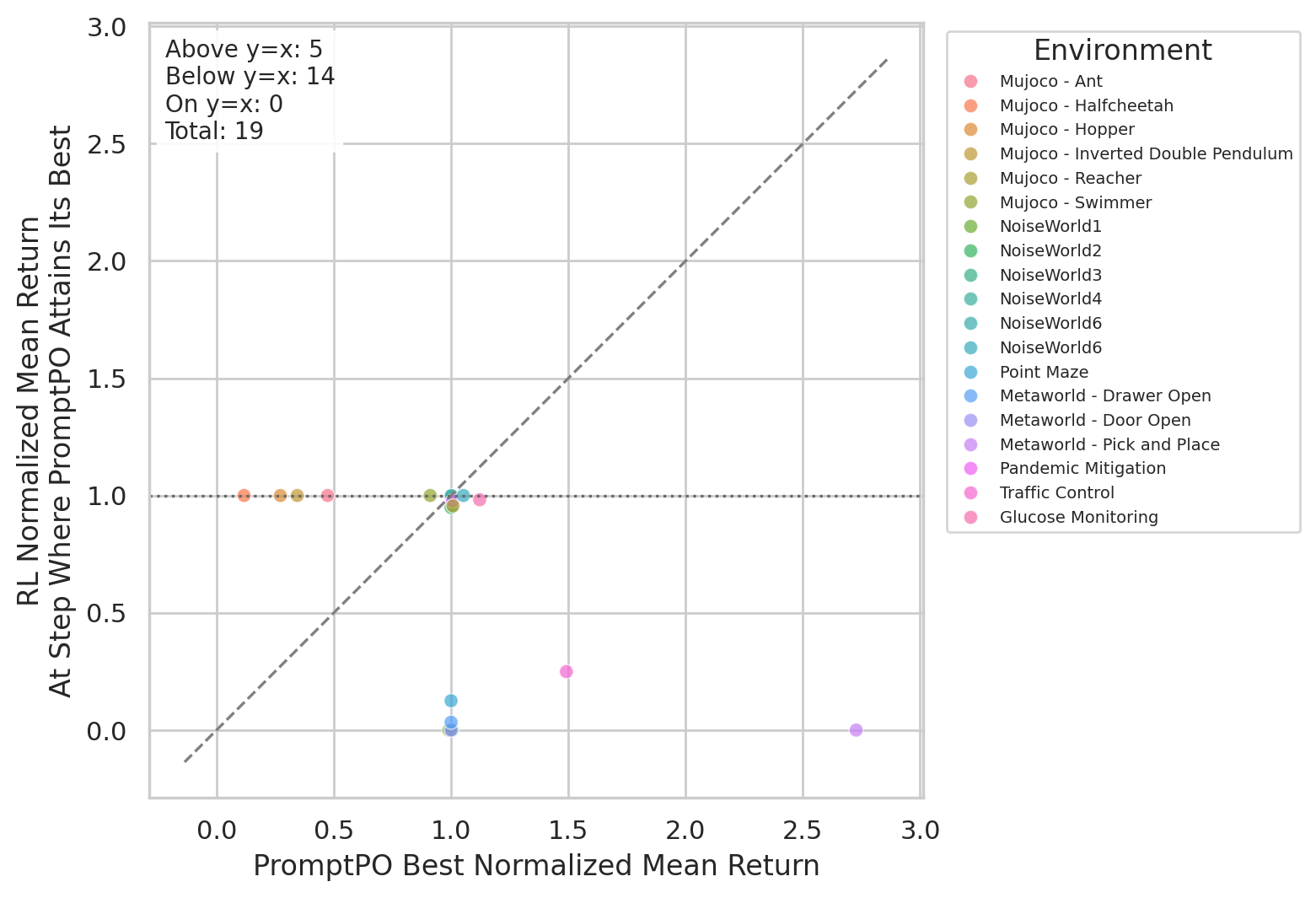}
        \caption{$N=20$}
        \label{fig:norm_mean_ret_summary_n20}
    \end{subfigure}
    \caption{
    PromptPO performance summary for different numbers of sampled candidate policies per round ($N \in \{5,10,20\}$).
    This figure complements Figure~\ref{fig:norm_mean_ret_summary}; see its caption for details on interpretation.
    }
    \label{fig:norm_mean_ret_summary_combined}
\end{figure}

The results show in Figures \ref{fig:main_fig} and \ref{fig:norm_mean_ret_summary}, as well as all other figures in the main text, sample $N=10$ candidate policy generations for each round of PromptPO. Figures \ref{fig:sample_efficiency_summary_combined}  and \ref{fig:norm_mean_ret_summary_combined} summarizes PromptPOs performance when $N \in \{5,10,20\}$, where the number of PromptPO update rounds remains fixed for all settings evaluated. Decreasing the number of candidate policies per round decreases the final performance attained; PromptPO benefits from sampling a diverse pool of candidates. Increasing the number of candidate policies per round decreases sample efficiency, as more policies need to be rolled out and evaluated in the environment.

\section{Providing exploration context for PromptPO}

\begin{figure}[H]
    \centering
    \includegraphics[width=0.6\linewidth]{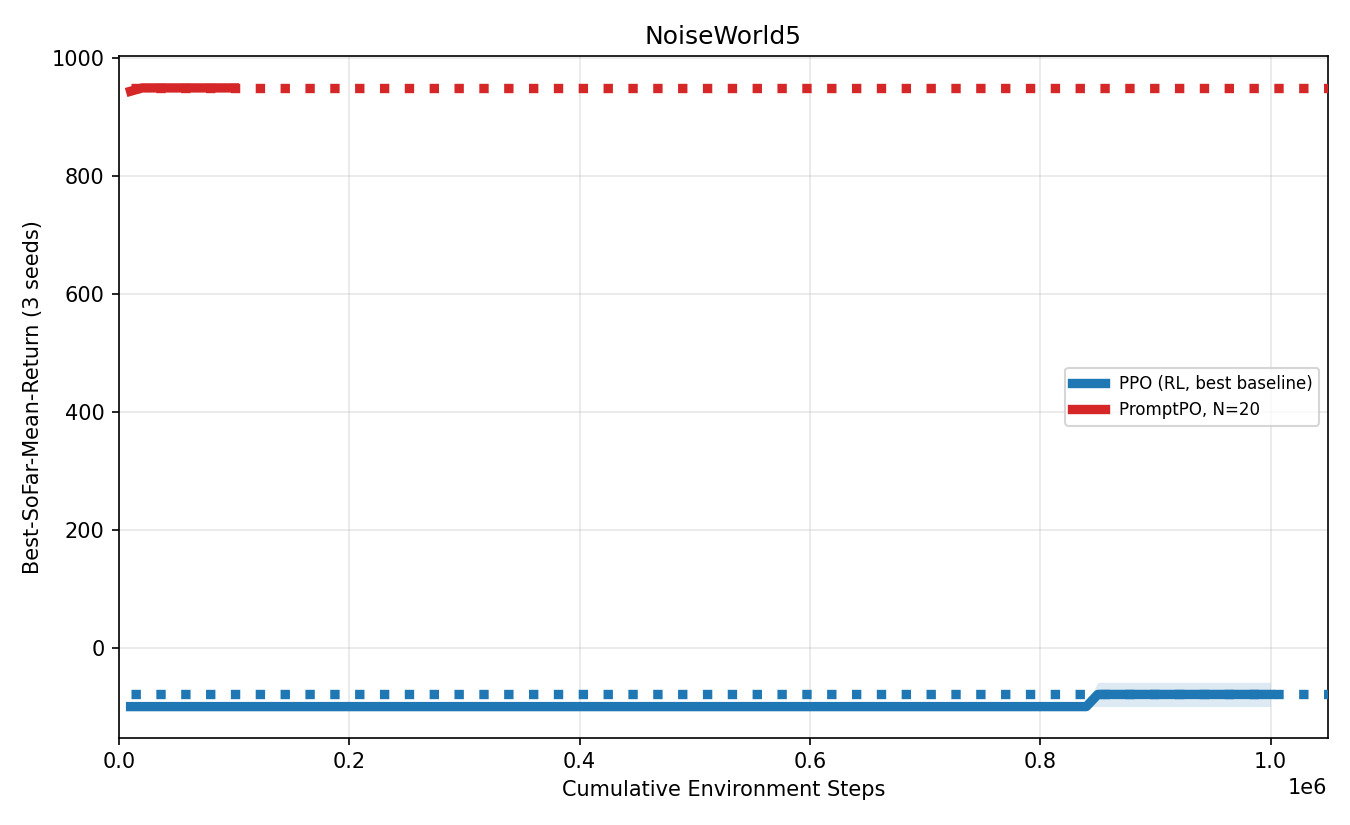}
    \caption{PromptPO's training performance in NoiseWorld5 versus PPO, which is the best performing RL algorithm out of the set of methods we consider. Here, PromptPO is provided with context stating it must visit cell 5 and cell 6 to reap the high positive reward at the goal state. Mean return is reported over 3 seeds. The dotted lines show best achieved final performance.}
    \label{fig:noiseworld5abalation}
\end{figure}

PromptPO, like the RL algorithms we consider, fail to find a near-optimal policy for NoiseWorld4 and NoiseWorld5, where a sequence of keys must be visited before the agent can attain high reward from the goal state. When providing the addition context in Figure \ref{fig:progress_flags_prompt}, however, PromptPO does attain near optimal performance as illustrated in \ref{fig:noiseworld5abalation}. This type of user-provided prior might be difficult to specify to an RL algorithm, but can be easily specified via natural language to PromptPO

\begin{figure}[H]
\centering
\begin{tcolorbox}[
    colback=gray!5,
    colframe=gray!50,
    boxrule=0.4pt,
    arc=2pt,
    left=4pt,
    right=4pt,
    top=4pt,
    bottom=4pt,
    title={Observation Context Addition},
    fonttitle=\bfseries,
    width=0.97\linewidth
]
\footnotesize
\ttfamily
Trailing progress flags (two scalars in \{0,1\}, after \texttt{board}):\\[0.4em]
First flag: 1 if the agent has already visited the cell labeled 6 at least once this episode, else 0.\\[0.4em]
Second flag: 1 if the agent has already visited the cell labeled 7 at least once this episode after the first flag became 1 (i.e.\ the ordered pair of milestones is complete), else 0.
\end{tcolorbox}
\caption{Additional observation-context text describing the two trailing progress flags appended after \texttt{board}.}
\label{fig:progress_flags_prompt}
\end{figure}

\end{document}